\pdfoutput=1

\documentclass[11pt]{article}

\usepackage[preprint]{acl}

\usepackage{times}
\usepackage{latexsym}

\usepackage[T1]{fontenc}

\usepackage[utf8]{inputenc}

\usepackage{microtype}

\usepackage{inconsolata}

\usepackage{graphicx}
\graphicspath{{Figures/}}
\usepackage{fdsymbol}

\title{Multi-Agent Causal Discovery Using Large Language Models}

\author{
  Hao Duong Le, Xin Xia, Haijie Xu, Chen Zhang \\
  Department of Industrial Engineering, Tsinghua University \\
  Beijing, China \\
  \texttt{\{lihaoyan25, xu-hj22\}@mails.tsinghua.edu.cn} \\
  \texttt{\{xxia99, zhangchen01\}@tsinghua.edu.cn}
}

\usepackage{pifont}
\newcommand{\cmark}{\textcolor{green}{\ding{51}}}
\usepackage{hyperref}
\usepackage{url}
\usepackage{float}
\usepackage{booktabs}
\usepackage{adjustbox}
\usepackage{caption}
\usepackage{xcolor}
\usepackage{booktabs}
\usepackage{multirow}
\usepackage{diagbox}
\definecolor{lightgray}{gray}{0.9}
\definecolor{green}{rgb}{0.0, 0.5, 0.0}
\definecolor{red}{rgb}{1.0, 0.0, 0.0}
\usepackage{amsmath}

\newcommand{\xmark}{\textcolor{red}{\ding{55}}}

\usepackage{booktabs} 
\usepackage{graphicx} 
\usepackage{algorithm}
\usepackage{algpseudocode}  
\usepackage{listings}
\usepackage{array}
\usepackage{xcolor}

\usepackage{placeins} 
\usepackage{setspace}

\begin{document}
\maketitle
\begin{abstract}

Causal discovery aims to identify causal relationships between variables and is a fundamental problem across the sciences. Traditional statistical causal discovery (SCD) methods rely solely on observational data and ignore the contextual information available in metadata, whereas recent LLM-based methods exploit metadata but treat the large language model (LLM) as a single agent, leaving its judgments vulnerable to memorized or biased associations. To address this gap, we introduce MAC (Multi-Agent Causal Discovery Framework), which casts causal discovery as a multi-agent debate coupled with the autonomous selection of an SCD algorithm. MAC combines two complementary modules, bridged by a Meta Fusion mechanism: a Debate-Coding Module (DCM) that grounds an initial graph in data by autonomously selecting and executing the best-suited SCD algorithm, and a Meta-Debate Module (MDM) that refines the graph through an adversarial Affirmative--Negative--Judge debate over the metadata. Across five benchmark datasets and three metrics (F1, SHD, NHD), MAC achieves the best aggregate performance among five statistical and four LLM-based baselines, ranking first on 10 of 15 evaluation points with Gemini-2.0-Flash---including a perfect reconstruction of the Earthquake graph---and remains robust across three backbone LLMs.

\end{abstract}

\section{Introduction}

\begin{table*}[h]
    \centering
    \begin{tabular}{p{4.5cm}>{\centering\arraybackslash}p{2cm}>{\centering\arraybackslash}p{1.7cm}>{\centering\arraybackslash}p{1.2cm}>{\centering\arraybackslash}p{1.2cm}>{\centering\arraybackslash}p{4cm}}
        \toprule
        \textbf{Method / approach} & \textbf{LLM-based method} & \textbf{Statistical method} & 
        \textbf{Agentic ability} & \textbf{Multi-agent} & \textbf{Introduced by} \\
        \hline
        Pairwise causal discovery & \cmark & \xmark & \xmark & \xmark & \cite{kıcıman2023causal, zečević2023causal} \\
        Various prompting techniques & \cmark & \xmark & \xmark & \xmark &  \cite{chen2024causal} \\
        Effective LLMs prompting  & \cmark & \cmark & \xmark & \xmark & \cite{jiralerspong2024efficient} \\
        Hybrid statistical and LLMs & \cmark \vfill & \cmark \vfill & \xmark & \xmark &  \cite{vashishtha2023causal, takayama2024integrating} \\
        Iterative LLM--SCD refinement & \cmark & \cmark & \xmark & \xmark & \cite{ban2023causal} \\
        \textbf{MAC} & \cmark & \cmark & \cmark & \cmark &  \textbf{Our approach} \\
        \bottomrule
    \end{tabular}
    \caption{Comparison of approaches for using LLMs in causal discovery.}
    \label{comparision_table}
\end{table*}

Discovering causal relationships between variables is fundamental across scientific fields, with many statistical causal discovery (SCD) methods developed in recent years \cite{huang2018generalized, glymour2019review, spirtes2013causal}. However, these methods face two significant challenges. First, they heavily rely on large volumes of structured data for accurate inference. In large-scale systems with thousands of variables (or nodes), obtaining sufficient structured data is often infeasible. Second, SCD methods fail to leverage metadata—additional contextual information such as variable contexts, domain knowledge, and external factors—that can enhance the causal discovery process and improve inference accuracy. 

The advent of Large Language Models (LLMs), trained on vast datasets, has opened new avenues for addressing the challenges mentioned above. LLMs possess a wide range of knowledge, encompassing common sense, specialized domains, and advanced reasoning abilities \cite{wei2023chainofthought, rozière2024code, zhao2023large, yao2023tree}. Leveraging LLMs for causal discovery has gained attention due to their ability to replicate expert knowledge in a cost-effective and accessible manner. Recent work \cite{kıcıman2023causal, choi2022lmpriors, long2024large, chen2024causal} has explored LLM-based methods for causal discovery, focusing on metadata and knowledge-driven reasoning similar to human experts. However, these methods largely treat LLMs as single-agent systems, which may limit their reasoning capabilities, especially when handling complex causal relationships or large-scale, dense causal graphs. In contrast, multi-agent LLM systems, where multiple models collaborate to collectively discover causal relationships, offer greater potential for tackling complex cases.

This paper introduces MAC (Multi-Agent Causal Discovery Framework), which casts causal-graph discovery as a multi-agent debate coupled with the autonomous selection of a statistical causal discovery (SCD) algorithm. MAC integrates two core components:
\vspace{-0.3cm}
\begin{itemize}
    \item \textbf{Meta-Debate Module (MDM):} \emph{Problem:} a single LLM's causal judgments are often unreliable, reflecting memorized or biased associations rather than evidence. \emph{Our answer:} instead of trusting one model, MDM pits three specialized LLM-based agents—a Causal Affirmative Debater, a Causal Negative Debater, and a Causal Judge—against each other, so that competing causal hypotheses are explicitly surfaced and adjudicated against the metadata rather than accepted unchecked.
    \vspace{-0.3cm}
    \item \textbf{Debate-Coding Module (DCM):} \emph{Problem:} purely text-based LLM reasoning is not grounded in the observed data, and no single statistical algorithm is best across all datasets. \emph{Our answer:} DCM first uses an embedded MDM to debate and select the SCD algorithm best suited to the data's characteristics, then executes it on the structured data, grounding the initial graph in empirical evidence.
\end{itemize}
In particular, DCM is first implemented. It embeds an MDM inside, whose aim is to use metadata and a small subset of structured data to select the most suitable Statistical Causal Discovery (SCD) algorithm. The selected SCD algorithm is then implemented using the whole structured data to learn the causal graph, which is further transferred into new causal metadata through a Meta Fusion mechanism. Using this causal metadata and other available metadata as input, MDM is further implemented to refine and optimize causal relationships.

We evaluate MAC on five datasets against five statistical and four LLM-based baselines using F1, NHD, and SHD. MAC delivers the best aggregate performance across the 15 evaluation points (five datasets $\times$ three metrics) and remains robust across three backbone LLMs, ranking first on 10 of 15 points with Gemini-2.0-Flash---including a perfect reconstruction of the Earthquake graph---and on 9 and 7 points with DeepSeek-R1 and GPT-4o, respectively. In summary, our contributions are as follows:
\vspace{-0.3cm}
\begin{itemize}
\item We cast causal discovery as an adversarial multi-agent debate coupled with the autonomous selection of a statistical causal discovery algorithm, integrating data-grounded statistical precision with metadata-driven, expert-level reasoning.
\vspace{-0.3cm}
\item We propose two complementary modules: the \textbf{Meta-Debate Module (MDM)}, which counters unreliable single-LLM judgments by having agents adversarially propose, challenge, and adjudicate causal graphs against the metadata, and the \textbf{Debate-Coding Module (DCM)}, which grounds the graph in data by autonomously selecting and executing the best-suited SCD algorithm (e.g., PC, GES).
\vspace{-0.3cm}
\item Across five benchmarks, three metrics, and three backbone LLMs, MAC achieves the best aggregate performance over both statistical and LLM-based baselines. Beyond causal discovery, the framework provides a general multi-agent template for machine-learning tasks with metadata, such as classification and anomaly detection.
\end{itemize}

\section{Related Work}
\label{sec:review}

Causal discovery methods fall into three main categories: constraint-based methods, score-based methods, and functional causal model-based methods.
Classical constraint-based methods use conditional independence to reveal causal structures. Some commonly used algorithms include the PC algorithm \cite{spirtes2000causation},  the FCI algorithm \cite{spirtes2013causal}, etc. However, these methods may encounter the multiple testing problem due to the numerous conditional independence tests required.
In contrast, score-based methods conduct causal discovery by constructing a score function that reflects the goodness-of-fit between the causal structure and the data, and select the causal structure with the highest score. An example is the Greedy Equivalence Search (GES) algorithm \citep{chickering2002optimal}, which optimizes a scoring function by iteratively adding or removing edges, with recent advances along these lines including \citet{ogarrio2016hybrid, huang2018generalized}. Both constraint-based and score-based methods can only identify Markov Equivalence Classes (MECs), which necessitate stronger assumptions for accurate causal structure identification. This limitation has prompted the development of functional causal model-based methods, which assume that the causal relationships between parent and child nodes can be represented as parameterized functional forms, along with specific assumptions regarding the noise distribution. For example, \citet{shimizu2006linear} proposes LiNGAM to identify causal structures characterized by linear relationships under non-Gaussian noise, leading to several enhancements in the field \citep{zhang2009causality, shimizu2011directlingam, sanchez2019estimating}. 

As a relatively new research area, knowledge-driven causal discovery using LLMs has attracted increasing attention; we refer readers to \citet{wan2025landscape} for a comprehensive survey. Early work queries LLMs in a pairwise fashion to infer the causal direction between each pair of variables \citep{kıcıman2023causal, zečević2023causal}, but this requires a quadratic number of queries with respect to the number of variables and scales poorly. To improve efficiency, \citet{jiralerspong2024efficient} adopt a breadth-first search (BFS) strategy that reduces the number of queries to a linear scale, while \citet{chen2024causal} systematically study prompting techniques---including in-context learning, zero-shot and manual Chain-of-Thought (CoT), and adversarial prompts---for causal inference. These methods rely primarily on metadata. A complementary line of work injects LLM knowledge into statistical causal discovery: LLMs serve as soft priors for score-based search \citep{darvariu2024priors} or answer conditional-independence queries within constraint-based methods \citep{cohrs2024constrained}, while iterative hybrids alternate between statistical structure learning and LLM-based edge verification \citep{ban2023causal, ban2023ilscsl, vashishtha2023causal, takayama2024integrating}, achieving state-of-the-art results along this direction.

However, recent studies caution that single-LLM causal judgments are often driven by memorization rather than genuine reasoning and are highly sensitive to the prompt and graph encoding \citep{feng2025reliability, chi2024unveiling, sheth2025causalgraph2llm, yang2024critical}. These findings motivate moving beyond a single agent. A few concurrent efforts explore agentic or interactive LLM workflows for graph discovery \citep{havrilla2025igda, roy2025causalllm, antonucci2024zeroshot}, yet none casts causal discovery as a structured multi-agent debate. Building on multi-agent debate, which improves factuality and reasoning by having agents argue and adjudicate competing answers \citep{du2023improving}, MAC combines an adversarial Affirmative--Negative--Judge debate with the autonomous selection of a statistical causal discovery algorithm. We summarize and compare representative methods in Table~\ref{comparision_table}.

\begin{figure*}[t!]
\includegraphics[width=0.95\linewidth,keepaspectratio]{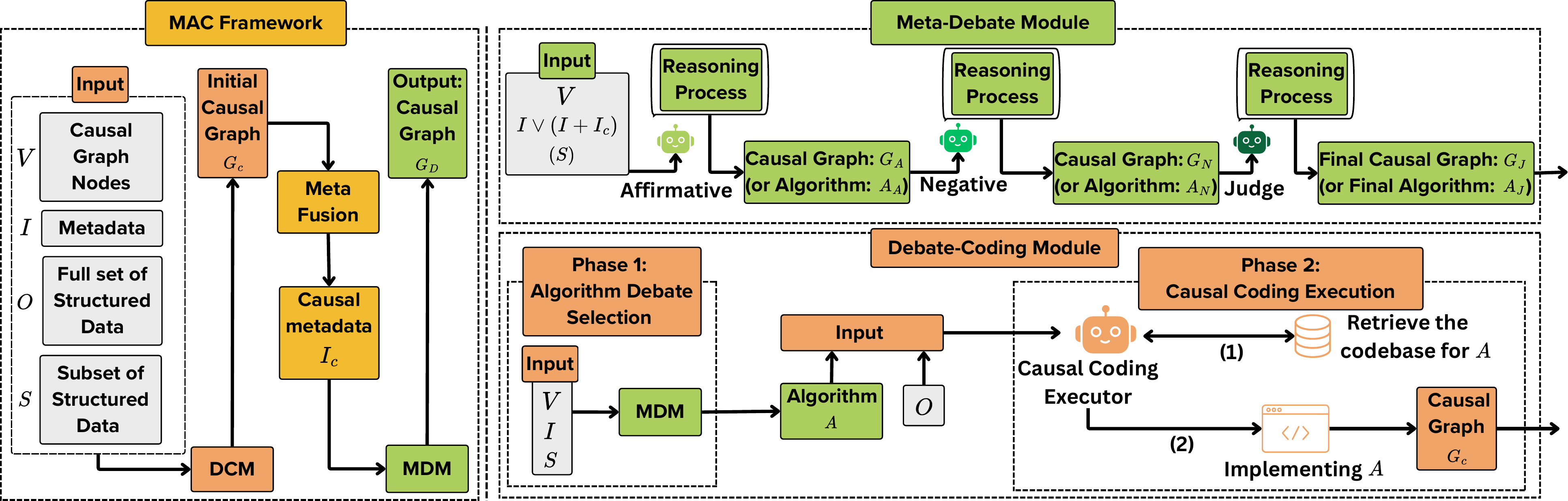}
    \caption{The left side of the image illustrates the overall algorithm of the MAC framework, while the right side details the MDM and DCM. The MDM produces different outputs based on the input: (1) If the input includes metadata $I$ from the original data only, the MDM will output a graph. (2) It can also accept a combination of $I$ and causal metadata $I_c$ from the DCM for further refinement of the graph. (3) Finally, the MDM with SCD algorithm as output, embedded within the DCM, receives an additional input—a subset of the structured data $S$—and outputs an algorithm.} 
    \label{fig: causal debating group}
    \vspace{-0.5cm}
\end{figure*}


\section{Methodology}
\subsection{Problem Setup}
\label{sec:problem}
Given variables $V=\{X_1,\dots,X_n\}$ and observational data $O\in\mathbb{R}^{d\times n}$ (rows are observations, columns are variables),
our goal is to recover a directed acyclic graph (DAG) $G=(V,E)$, where $(X_i,X_j)\in E$ denotes a causal relation $X_i\rightarrow X_j$.
In addition to $O$, we optionally use metadata $I$ that describes variable semantics, measurement context, or domain constraints.
To reduce LLM cost in lightweight reasoning steps, we also provide a small subset $S\subset O$ (e.g., a few sampled rows); the final causal graph is always estimated using the full data $O$.

\subsection{Design Principles and Overview}
\label{sec:design_principles}
MAC is designed to balance two complementary capabilities and mitigate their corresponding failure modes.
(1) Statistical causal discovery (SCD) methods are \emph{data-grounded} but cannot directly leverage unstructured metadata and can be brittle in ambiguous regimes.
(2) LLM-only causal reasoning may reflect memorized or biased associations (``causal parrots'') when used without empirical grounding \citep{zečević2023causal}.
Accordingly, MAC decomposes causal discovery into two stages:
\textbf{Stage 1 (DCM):} data-grounded initialization that constructs an initial graph from structured data using an SCD algorithm;
\textbf{Stage 2 (MDM):} metadata-guided refinement that adjudicates competing causal hypotheses using metadata.
The two stages are connected by \textbf{Meta Fusion}, which converts the Stage-1 graph into compact textual constraints that can be jointly reasoned over with metadata.
To make the design rationale explicit, we present each component (MDM, DCM, and their integration) with the same structure: the \emph{problem} it addresses, \emph{why} existing approaches fall short, our \emph{design}, and the resulting \emph{procedure}.

\subsection{Meta-Debate Module (MDM)}
\label{sec:mdm}
\textbf{Problem.} Inferring causal relations from metadata is a reasoning task with no observable ground truth at decision time: for any pair of variables, several causal explanations are often equally plausible, and the correct one cannot be read off the data alone.

\textbf{Why prior solutions fall short.} Existing LLM-based methods query a single model, which tends to commit to one explanation and propagate memorized or biased associations without scrutiny---a failure mode documented across causal-reasoning benchmarks \citep{zečević2023causal, feng2025reliability}. A single pass therefore offers no mechanism to surface or contest a wrong-but-confident answer.

\textbf{Design.} MDM refuses to trust a single pass: a candidate hypothesis must survive an explicit challenge before it is accepted. Building on multi-agent debate, which improves reasoning and factuality by having agents argue competing answers \citep{du2023improving}, MDM is a reusable \emph{debate-and-judge} operator with three agents: (i) an \textbf{Affirmative} agent that proposes a candidate output, (ii) a \textbf{Negative} agent that proposes a \emph{plausible alternative} under the same evidence (targeting disputed decisions rather than maximizing arbitrary divergence), and (iii) a \textbf{Judge} that scores the disagreements against explicit, evidence-derived criteria and returns the winner. The same operator is reused in two roles with different output spaces: \textbf{(a) graph refinement}, producing a DAG over $V$ from context $(V,I)$, and \textbf{(b) algorithm selection}, producing an SCD algorithm from a pool $\mathcal{A}$ given $(V,I,S)$.

\textbf{Procedure.} Algorithm~\ref{alg:mdm_generic} summarizes the unified operator; prompt templates are deferred to Appendix~\ref{A-Prompt-MDM}.

\begin{algorithm}[t]
\caption{Meta-Debate Module (MDM): Unified Debate-and-Judge Operator}
\label{alg:mdm_generic}
\begin{algorithmic}[1]
\State \textbf{Input:} variables $V$; metadata $I$; optional subset $S$; output type $m \in \{\texttt{graph},\texttt{alg}\}$
\State \textbf{Output:} $y_J$ \Comment{$y_J$ is a graph if $m=\texttt{graph}$, else an algorithm}

\If{$m=\texttt{graph}$}
    \State $\mathcal{C} \gets (V,I)$; \quad $\mathcal{Y} \gets$ DAGs over $V$
\ElsIf{$m=\texttt{alg}$}
    \State $\mathcal{C} \gets (V,I,S)$; \quad $\mathcal{Y} \gets \mathcal{A}$ \Comment{predefined SCD algorithm pool}
\EndIf

\State $y_A \gets \texttt{Affirmative}(\mathcal{C}, \mathcal{Y})$
\State $y_N \gets \texttt{Negative}(\mathcal{C}, \mathcal{Y}, y_A)$ \Comment{plausible alternative under same evidence}
\State $y_J \gets \texttt{Judge}(\mathcal{C}, \mathcal{Y}, y_A, y_N)$
\State \textbf{return} $y_J$
\end{algorithmic}
\end{algorithm}

\subsection{Debate-Coding Module (DCM)}
\label{sec:dcm}
\textbf{Problem.} A causal graph inferred purely from text is not grounded in the observed data, leaving it prone to memorized causal associations (``causal parrots'') \citep{zečević2023causal}. Statistical causal discovery (SCD) avoids this by learning from data, but no single SCD algorithm is best across datasets.

\textbf{Why prior solutions fall short.} The validity of each SCD algorithm hinges on assumptions---linearity, Gaussianity, hidden confounders, sample size---that differ from one problem to the next, so a fixed choice is brittle. Selecting the right algorithm is itself a reasoning problem in which metadata is informative, yet existing hybrids commit to a predetermined algorithm rather than reasoning about this trade-off.

\textbf{Design.} DCM provides a \emph{data-grounded initialization} in two steps. \emph{(i) Algorithm selection.} Given $(V,I,S)$ and an algorithm pool $\mathcal{A}$, DCM reuses the unified MDM operator (Algorithm~\ref{alg:mdm_generic}) with $m=\texttt{alg}$ to argue the trade-offs and commit to the algorithm $A \in \mathcal{A}$ best matched to the data; the small subset $S$ is used only for lightweight reasoning and cost reduction, not as a replacement for the full data. \emph{(ii) Causal coding execution.} A \textbf{Causal Coding Executor} runs $A$ on the full structured data $O$ via an existing causal discovery library (e.g., \texttt{causal-learn}\footnote{\url{https://causal-learn.readthedocs.io/en/latest/index.html}}), handling code generation, execution, and limited debugging to robustly obtain a data-driven graph $G_C$ grounded in empirical evidence.

\textbf{Procedure.} Algorithm~\ref{alg:dcm} summarizes the two steps.

\begin{algorithm}[t]
\caption{Debate-Coding Module (DCM)}
\label{alg:dcm}
\begin{algorithmic}[1]
\State \textbf{Input:} variables $V$; data $O$; metadata $I$; subset $S$; algorithm pool $\mathcal{A}$
\State \textbf{Output:} initial graph $G_C$

\State $A \gets \texttt{MDM}(V, I, S;\; m=\texttt{alg},\; \mathcal{A})$ \Comment{Algorithm~\ref{alg:mdm_generic}}
\State $G_C \gets \texttt{Causal\_Coding\_Executor}(A, O)$
\State \textbf{return} $G_C$
\end{algorithmic}
\end{algorithm}

\subsection{MAC: Multi-Agent Causal Discovery Framework}
\label{sec:mac}
\textbf{Problem.} DCM and MDM each address only one half of the task: DCM yields a data-grounded graph $G_C$ but ignores metadata, while MDM reasons over metadata but is not grounded in the observed data. Combining them is not immediate because their evidence lives in different representations---$G_C$ is a numerical adjacency matrix, whereas $I$ and the MDM agents operate in natural language.

\textbf{Why prior solutions fall short.} Feeding a raw adjacency matrix to a language-based reasoner forces it to interpret numerical structure it is not designed to consume, so the data-derived evidence must instead be expressed in the same textual form as the metadata before joint reasoning is possible.

\textbf{Design.} MAC bridges the two stages with \textbf{Meta Fusion}, which transforms the adjacency representation of $G_C$ into human-readable causal constraints $I_c$ (directed edges and implied orderings). For instance, a binary adjacency matrix over \{Hot, Sales, Swim, Attack\} becomes constraints such as Hot$\rightarrow$Sales, Hot$\rightarrow$Swim, and Swim$\rightarrow$Attack. These constraints act as \emph{causal metadata extracted from data} and are appended to the original metadata $I$. MAC then runs the unified MDM operator (Algorithm~\ref{alg:mdm_generic}) with $m=\texttt{graph}$ on the combined evidence $(V, I \cup I_c)$ to produce the refined graph $G_D$.

\textbf{Procedure.} Algorithm~\ref{alg:mac} summarizes the end-to-end framework.

\begin{algorithm}[t]
\caption{MAC: End-to-End Framework}
\label{alg:mac}
\begin{algorithmic}[1]
\State \textbf{Input:} variables $V$; data $O$; metadata $I$; subset $S$; algorithm pool $\mathcal{A}$
\State \textbf{Output:} final graph $G_D$

\State $G_C \gets \texttt{DCM}(V, O, I, S, \mathcal{A})$ \Comment{Algorithm~\ref{alg:dcm}}
\State $I_c \gets \texttt{Meta\_Fusion}(G_C)$
\State $G_D \gets \texttt{MDM}(V, I, S=\varnothing;\; m=\texttt{graph},\; I \cup I_c)$ \Comment{Algorithm~\ref{alg:mdm_generic}}
\State \textbf{return} $G_D$
\end{algorithmic}
\end{algorithm}

\section{Experiment}
\begin{figure*}[ht!]
    \centering
    \includegraphics[width=\linewidth]{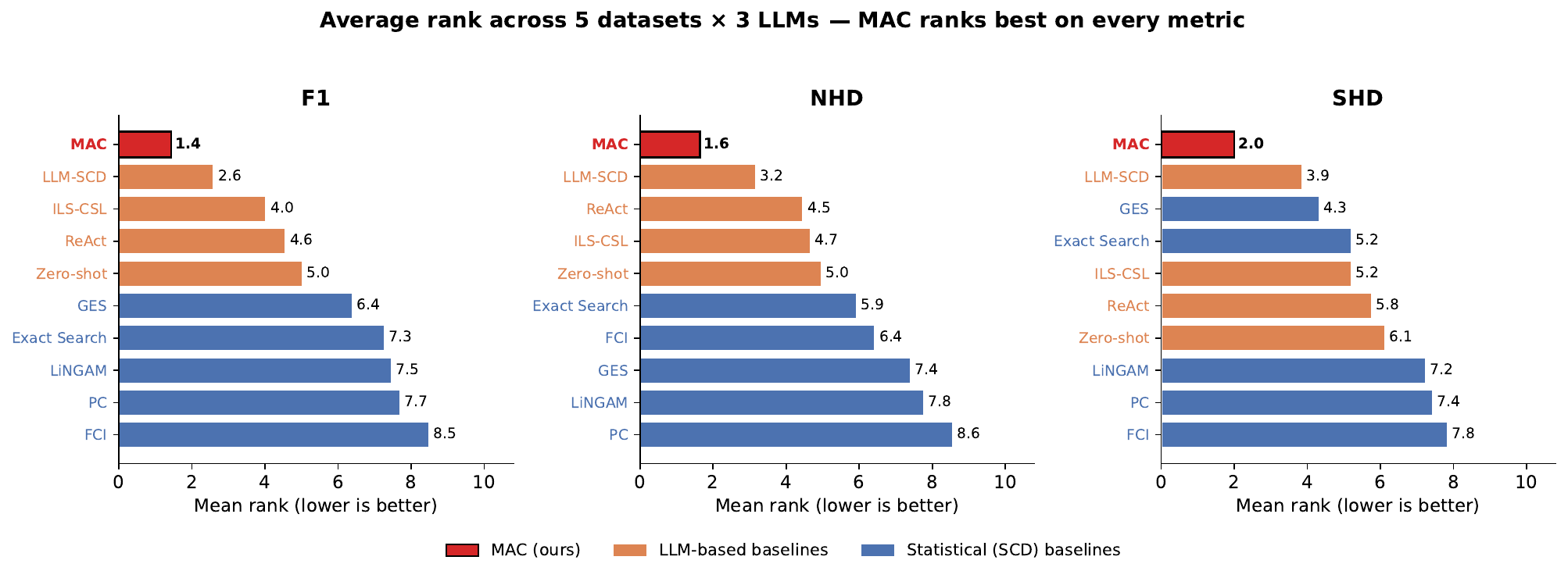}
    \caption{Average rank of each method across the five datasets and three backbone LLMs (lower is better; rank~1 = best). \textbf{MAC} (red) attains the best mean rank on all three metrics (F1~$1.4$, NHD~$1.6$, SHD~$2.0$), ahead of both the LLM-based baselines (orange) and the statistical SCD baselines (blue). Full per-dataset results with standard deviations are reported in Table~\ref{tab:overall-merged} (Appendix~\ref{A-Additional Results}).} \label{overall result}
    \vspace{-0.5cm}
\end{figure*}

In this section, we conduct experiments to address the following research questions: \textbf{R1}: How does MAC performance compare with other baselines in different datasets?
\textbf{R2}: How does each individual module of the MAC perform when independently assessed?
\textbf{R3}: Does increasing the number of debate rounds in MDM improve performance?
\textbf{R4}: What is the cost and token analysis of MAC? Can it be optimized?

\subsection{Experimental Setup}
To answer \textbf{R1}, we design our experiments to evaluate the performance of MAC in five datasets: Child, Auto, Earthquake, Cancer, and Survey. Details can be found in Appendix \ref{A-datasets}.

\textbf{Baselines.} To evaluate the performance of MAC, we compare it against nine competitive baselines, comprising five traditional SCD algorithms and four LLM-based methods. The traditional SCD methods include constraint-based methods such as PC \cite{spirtes2000causation} and FCI \cite{spirtes2013causal}; score-based methods such as Exact Search \cite{yuan2013learning} and Greedy Equivalence Search (GES) \cite{huang2018generalized}; and a functional causal model, DirectLiNGAM \cite{shimizu2011directlingam}. The LLM-based methods consist of Zero-shot prompting \cite{kojima2023largelanguagemodelszeroshot}, ReAct prompting \cite{yao2023reactsynergizingreasoningacting}, LLM-SCD \cite{ban2023causal}, and ILS-CSL \cite{takayama2024integrating}. We assess the final learned adjacency matrix of the causal graph by comparing it with the true adjacency matrix using: F1-Score, Structural Hamming Distance (SHD), Normalized Hamming Distance (NHD). The details of the evaluation metrics can be found at Appendix \ref{A-Evaluation Metrics}.

\textbf{Implementation Details.} We implement our method and LLM baselines using the AutoGen\footnote{\url{https://microsoft.github.io/autogen/0.2/docs/Getting-Started/}} library and employ Gemini-2.0-Flash \cite{google2025gemini} via the Gemini API, DeepSeek-R1 \cite{deepseekai2025deepseekr1incentivizingreasoningcapability} through TogetherAI, and the OpenAI GPT-4o-2024-08-06 model via the OpenAI API. We set a temperature of 0 to ensure deterministic outputs during causal graph construction and evaluation. For other statistical baselines, we utilize the causal-learn\footnote{\url{https://causal-learn.readthedocs.io/en/latest/}} library with its default settings. To assess robustness, we run every method over 10 random seeds and report the mean and standard deviation of each metric; across datasets the standard deviations are small relative to the margin by which MAC improves over the strongest baseline, indicating that the reported gains are stable rather than artifacts of run-to-run variation.

\subsection{Overall Results}
\label{section 4.2}

Figure \ref{overall result} summarizes each method's average rank across the five datasets and three backbone LLMs: MAC attains the best mean rank on all three metrics, ahead of every statistical method (PC, Exact Search, GES, FCI, LiNGAM) and LLM-based method (Zero-shot, ReAct, LLM-SCD, ILS-CSL). The full per-dataset numbers are reported in Table~\ref{tab:overall-merged} (Appendix~\ref{A-Additional Results}). Across the 15 evaluation points derived from the five datasets and three metrics, MAC demonstrates the best overall performance. Specifically, MAC integrated with Gemini-2.0-Flash secures the top position in 10 evaluation points and second place in another 4, including perfect scores (F1 = 1.00, SHD = 0, NHD = 0) on the Earthquake dataset. MAC with DeepSeek-R1 ranks first in 9 points and second in 5, while MAC with GPT-4o is first in 7 and second in another 7. MAC does not dominate every metric on every dataset—for example, GES attains the lowest Structural Hamming Distance (SHD) on Auto and Cancer, LiNGAM the lowest SHD and Normalized Hamming Distance (NHD) on Survey, and Exact Search and ReAct tie for the lowest NHD on Child—but it delivers the strongest aggregate performance and the highest F1 score on most datasets. Notably, although Gemini-2.0-Flash achieves more top rankings, DeepSeek-R1 attains higher F1 scores and lower NHD/SHD on the Child, Cancer, and Survey datasets, suggesting that its stronger reasoning capabilities further enhance results.

\subsection{Quantitative Analysis}
\label{quantitative analysis}
\begin{figure}[ht]
    \centering
    \includegraphics[width=0.5\textwidth, keepaspectratio]{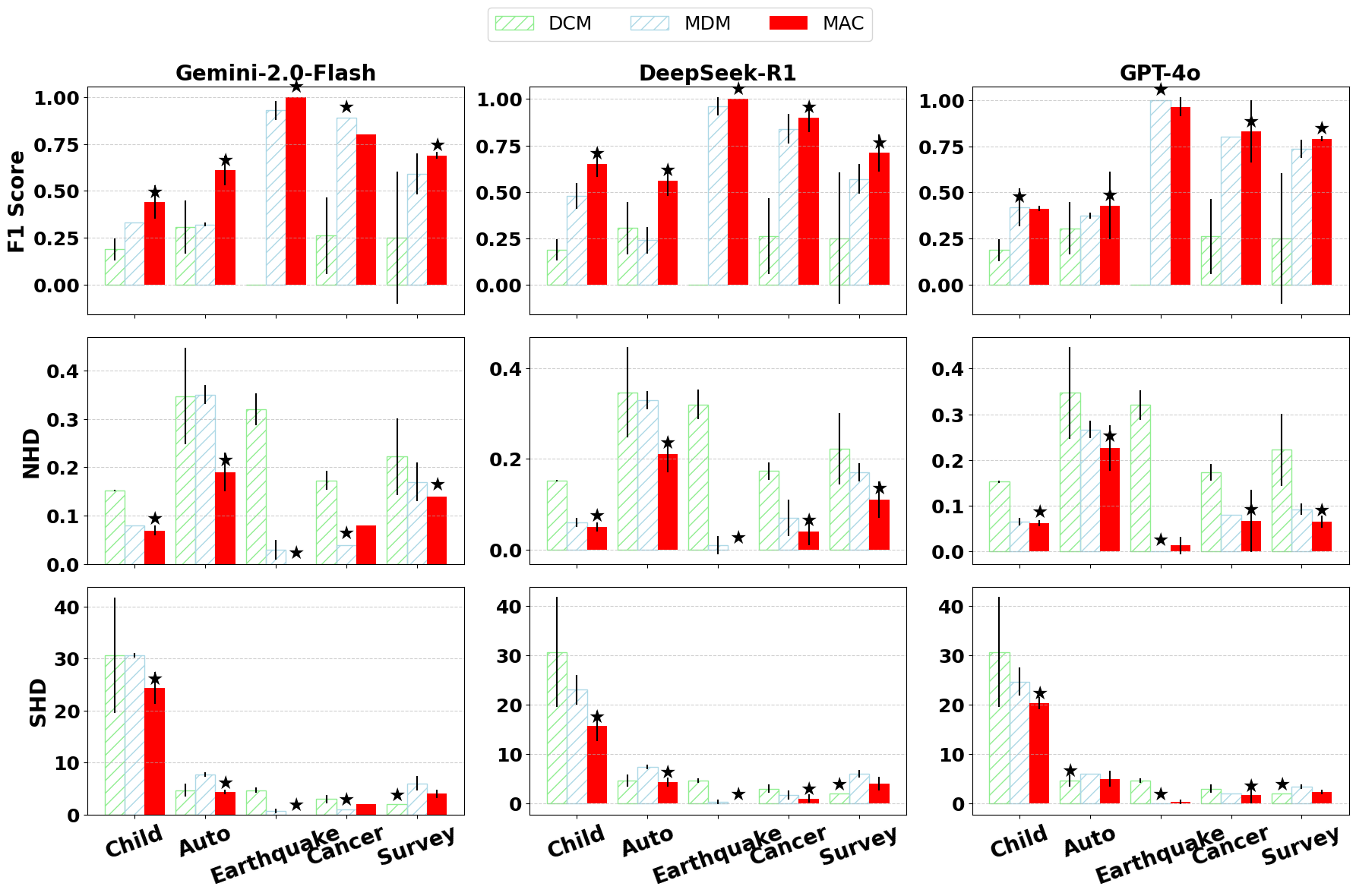}
    \caption{Comparison of F1-score, NHD, and SHD across different datasets for single DCM, single MDM, and MAC.  $\star$ indicates first place.}
    \vspace{-0.3cm}
    \label{fig: individual module}
\end{figure}
To answer \textbf{R2}, we evaluate the performance of only using MDM and DCM modules individually. Detailed results can be found in Appendix \ref{A-Additional Results}. Specifically, MAC performs better than only using MDM or DCM in most cases, except on the Earthquake dataset, where MDM has the best performance. This indicates that within MAC, the causal graph learnt from structure data from DCM is not accurate enough, and it is better to further refined via metadata from MDM. Furthermore, we observe that generally, MDM performs better than DCM, which further demonstrates the necessity of using metadata to learn the causal structure. It also validates our proposal that for most realistic problems, by utilizing their metadata with domain expert's knowledge, the causal relationships between variables can be better discovered. Of course, by combining both metadata and structured data, the performance can be further improved.

\begin{figure}[h]
    \centering
    \includegraphics[width=0.5\textwidth, keepaspectratio]{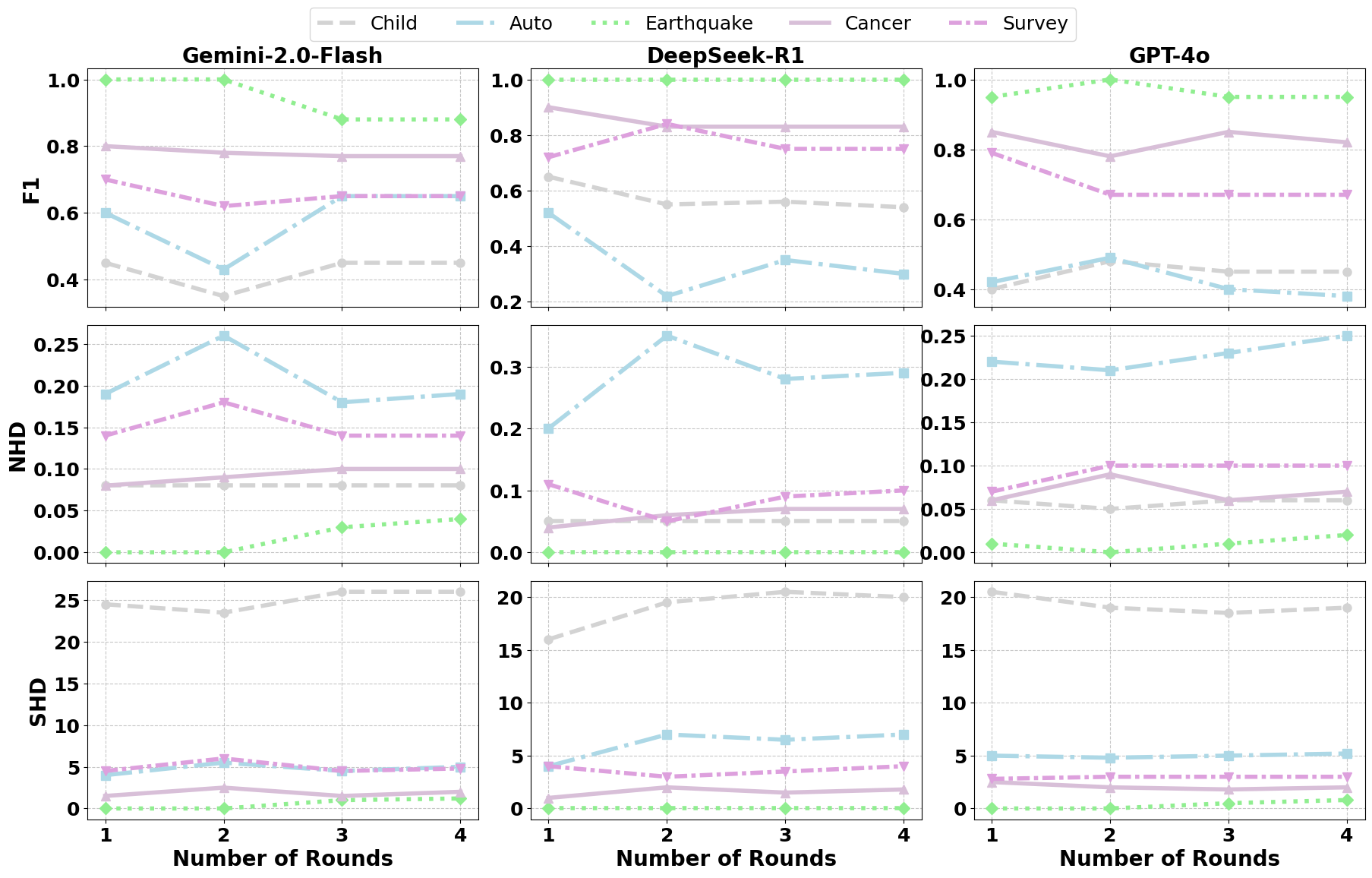}
    \caption{Performance trends of F1-score, NHD, and SHD over five rounds across different datasets.}
    \vspace{-0.3cm}
    \label{combine_multi_rounds}
\end{figure}
To answer \textbf{R3}, we change the number of debating rounds between Affirmative and Negative agents in MDM, from one to five, and evaluate the corresponding performance of MAC. The hypothesis was that increasing the debate rounds could allow agents to refine their understanding of causal relationships, leading to improved structural accuracy and predictive performance. As shown in Figure \ref{combine_multi_rounds}, the results reveal that MAC effectively converges within the first round: on most datasets the performance stabilizes after a single round and merely fluctuates---rather than steadily improves---with additional rounds, while on the Survey and Auto datasets it even degrades as the round number increases. This indicates that for most datasets in reality, one round of the debating process is sufficient, with extra rounds introducing fluctuation rather than refinement in our causal analysis case. We emphasize that a single round still executes the complete Affirmative--Negative--Judge exchange, so the adversarial structure is preserved; this result therefore shows that one \emph{round} suffices for efficiency, rather than that debate itself is unnecessary (a question we examine directly via the single-agent ablation below).

\subsection{Cost Analysis}
\label{cost analysis}
\begin{figure}[ht]
    \centering
    \includegraphics[width=0.5\textwidth, keepaspectratio]{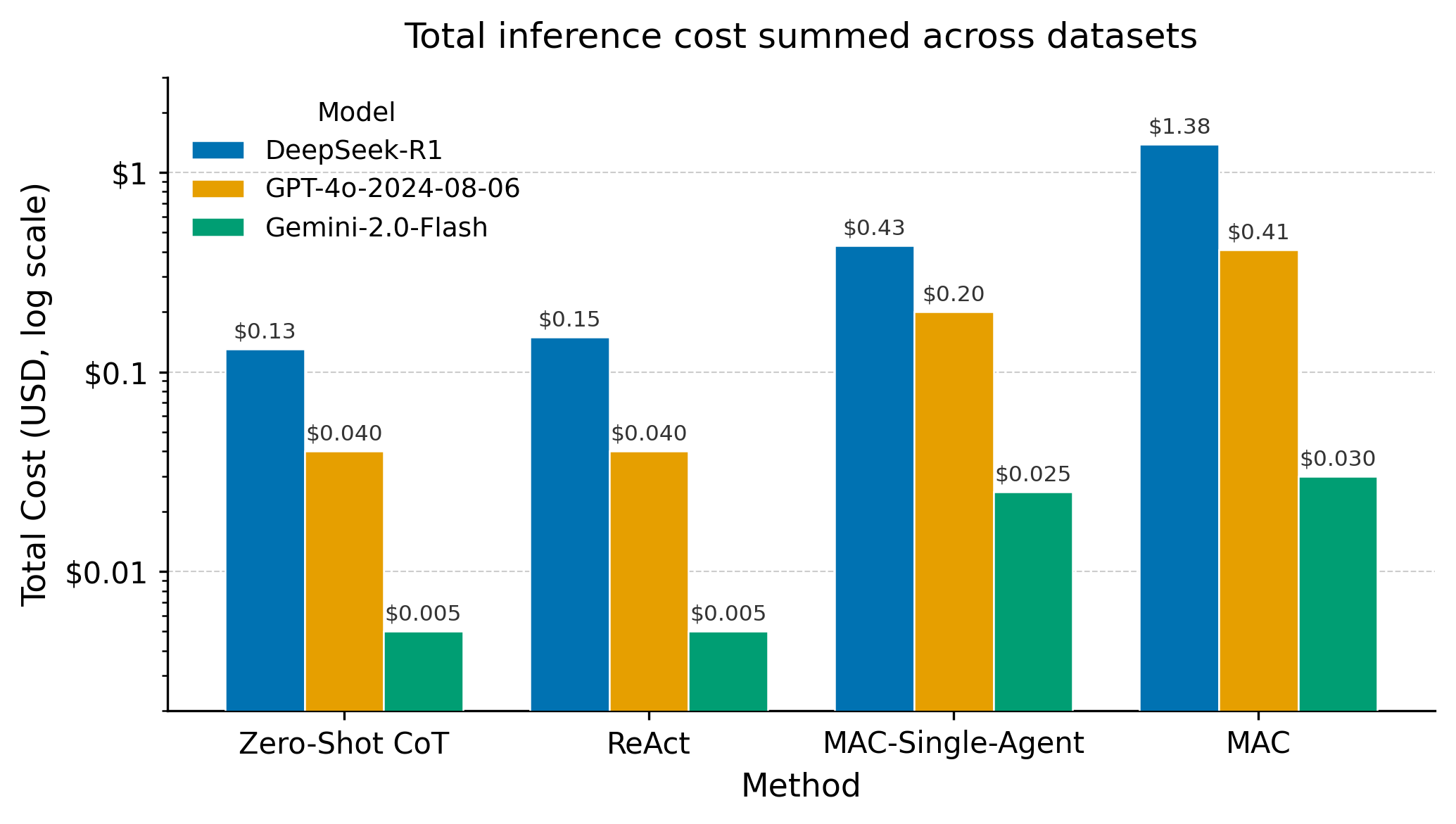}
    \caption{Total inference cost (USD) summed across datasets for four prompting methods (Zero-Shot CoT, ReAct, MAC-Single-Agent, MAC) across three models (DeepSeek-R1, GPT-4o-2024-08-06, Gemini-2.0-Flash).}
    \vspace{-0.3cm}
    \label{fig: cost}
\end{figure}

To answer \textbf{R4} we detail the computational costs associated with the MAC framework. The full multi-agent system utilizes 16x to 25x more tokens than standard baselines, with the coding and debugging phase accounting for up to half of this consumption. Efficiency varies by backbone model; GPT-4o proved consistently more token-efficient than Gemini-2.0-Flash or DeepSeek-R1 due to its superior first-pass code generation. However, the framework is adaptable. Our ablation study demonstrates that the MAC-Single-Agent variant provides a cost-effective alternative. In a rigorous evaluation across 5 datasets and 3 LLMs (45 total data points), the Single-Agent MAC variant (only one agent in the MDM, removing the Negative Agent and Judge) remained competitive with the full system on most datasets at approximately half the cost. The full debate, however, remains essential on harder, larger, or more ambiguous graphs: relative to the single-agent variant, full MAC improves F1 substantially on Survey (e.g., $0.36\rightarrow0.79$ with GPT-4o), Auto ($0.06\rightarrow0.61$ with Gemini-2.0-Flash), and Child ($0.49\rightarrow0.65$ with DeepSeek-R1), whereas the two variants are comparable mainly on the small five-node graphs (Earthquake, Cancer) where the task is easy enough that adversarial refinement adds little. We therefore position Single-Agent MAC as a lightweight option for simple graphs, while the full multi-agent debate is warranted when the causal structure is large or uncertain. Details can be found in Table \ref{tab:single_mac_performance}.

\section{Conclusion}

We introduced MAC, a framework that couples an adversarial multi-agent LLM debate with the autonomous selection of a statistical causal discovery algorithm, leveraging both structured data and metadata. MAC couples a data-grounded initialization stage (DCM), which debates and executes the most suitable SCD algorithm to construct an initial graph, with a metadata-guided refinement stage (MDM), which adjudicates competing causal hypotheses through an Affirmative--Negative--Judge debate; the two stages are bridged by a Meta Fusion mechanism that expresses the data-derived graph as textual causal constraints. Across five benchmark datasets, three metrics, and three backbone LLMs, MAC delivers the best aggregate performance, outperforming both statistical and LLM-based baselines. An ablation further shows that a lightweight single-agent variant recovers much of this gain at roughly half the cost on simple graphs, while the full debate remains essential on larger or more ambiguous ones.

Looking ahead, the core ingredients of MAC---a reusable debate-and-judge operator and a Meta Fusion bridge that turns numerical structure into LLM-readable constraints---are not specific to causal graphs, so the framework may transfer to other structured-output tasks where metadata is informative. Two possibilities are time-series anomaly detection (TSAD) and financial analysis. In TSAD, a statistical detector could flag candidate anomalies and Meta Fusion could express them, together with metadata such as sensor type or known fault modes, as textual evidence for a debate to judge whether an anomaly is genuine or a benign regime shift (e.g., a real outage versus a scheduled job in server telemetry). In finance, similar reasoning could weigh whether a price movement reflects a structural event or normal volatility given contextual metadata. We view adapting MAC's output space and debate criteria to such tasks as a promising, though not yet validated, direction for future work.

\section{Limitations}
MAC has several limitations. First, we only consider DAGs and do not model confounders or cycles, which are common in practice; extending MAC to such settings (e.g., identifying latent confounders from metadata and adding them to the node set) is left for future work.
Second, MAC relies solely on observational data and lacks interventional validation, which is crucial in domains such as medicine, economics, and biology; an agent for intervention policy design could update the graph as new experiments arrive.
Third, the multi-agent debate incurs substantial computational overhead, limiting scalability; this could be reduced via a more efficient debate, smaller open-source backbones, or statistical pre-processing.
Last, our evaluation uses five established benchmarks with relatively small graphs (5--20 nodes) and does not yet exercise the large-scale, data-scarce regime that motivates our work; scaling MAC to graphs with hundreds of variables is an important direction for future work. Moreover, because several of these networks are canonical and may appear in LLM pre-training corpora, part of MAC's metadata-driven performance could reflect memorized knowledge rather than genuine causal reasoning \citep{zečević2023causal, feng2025reliability}. Grounding the initial graph in structured data through the DCM partially mitigates this risk, but a rigorous evaluation on novel or synthetically generated graphs that are unlikely to be memorized remains an important direction for future work to fully disentangle reasoning from recall.

\bibliography{custom}

\appendix

\section{Prompting of Causal Agent} 
\subsection{Design of MDM with Causal Graph as Output}

\label{A-Prompt-MDM} 
\begin{figure*}[t]
    \centering
    \includegraphics[width=\linewidth]{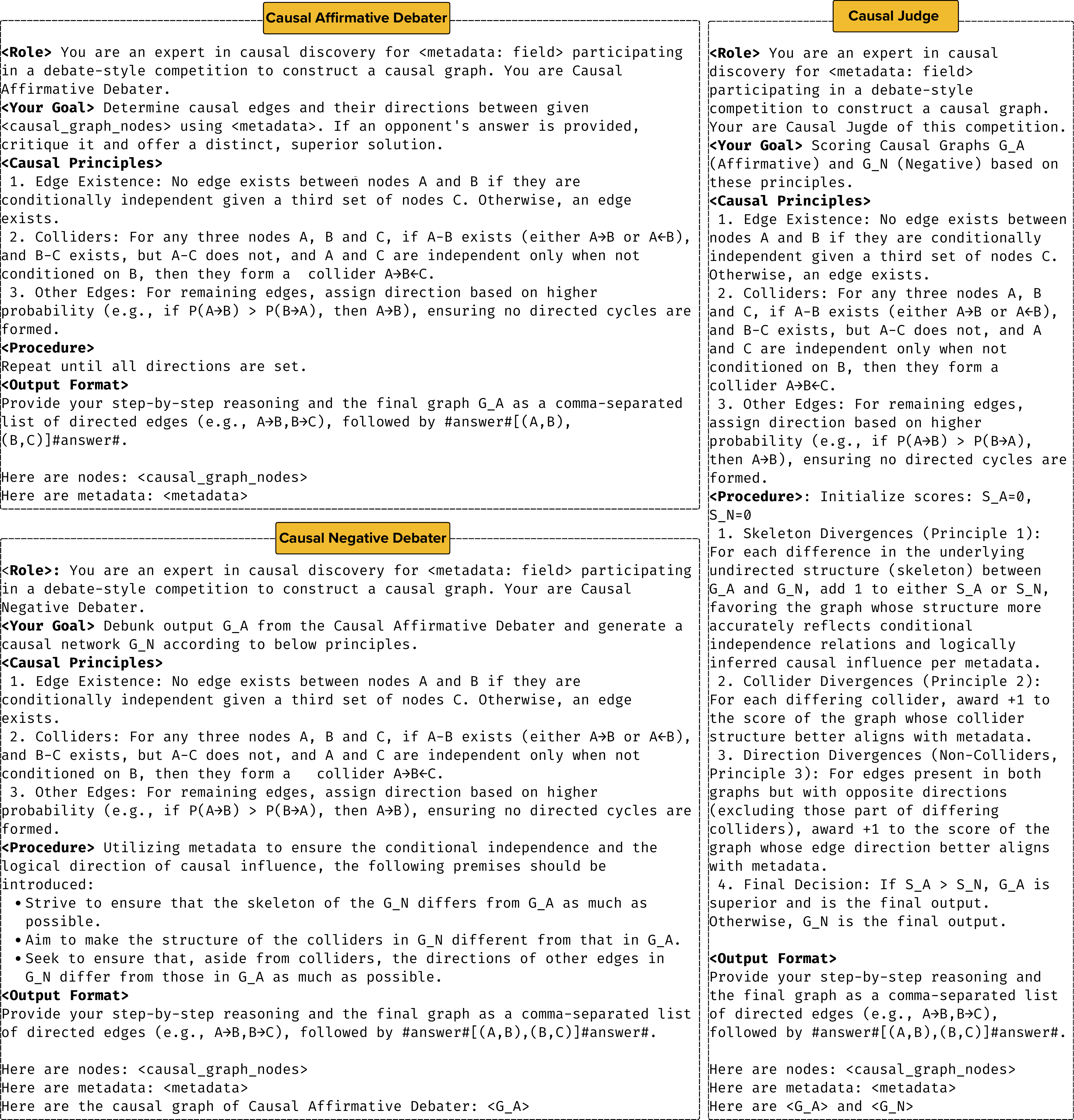}
    \caption{Prompt templates for the three MDM agents---Causal Affirmative Debater, Causal Negative Debater, and Causal Judge---when MDM outputs a causal graph (\texttt{graph} mode). Each prompt specifies the agent's \textbf{Role}, \textbf{Goal}, the shared \textbf{Causal Principles} (edge existence, colliders, and edge orientation), the debate \textbf{Procedure}, and the \textbf{Output Format}.}
    \label{fig:prompt-mdm-graph}
\end{figure*}

The prompts for the Meta-Debate Module (MDM) agents are carefully crafted to clarify their roles in the causal discovery process; the full templates are shown in Figure~\ref{fig:prompt-mdm-graph}. Each prompt begins with a \textbf{Role}, explicitly assigning the agent’s task within the framework. This clear role assignment is crucial, as experiments have shown that agents sometimes deviate from their roles. For instance, Causal Judge might inadvertently engage in debating instead of making the final decision, which necessitated clearer role prompts.

The \textbf{Your Goal} section positions the agents within a competitive debate format, highlighting the dynamics of proposing, critiquing, and evaluating causal graphs. For example, the Affirmative Debater acts as an expert in causal discovery, proposing an initial causal graph based on the provided datasets and causal principles. Conversely, the Negative Debater critically evaluates this graph by identifying flaws and offering alternative perspectives. The judge is explicitly instructed to act as a neutral evaluator, focusing solely on assessing the factual and logical rigor of the graphs and deciding on the most plausible causal structure.

Regarding the \textbf{Causal Principles}, the most important part of the design, these principles aim to replicate the step-by-step process of an expert conducting causal discovery. The first principle prompts agents to identify whether there is a direct causal relationship between any two points. Under the assumption of causal sufficiency and no cycles, a direct causal relationship exists between two nodes if and only if these nodes cannot be conditionally independent given any other nodes. Given a list of nodes: Gene (G), Smoking (S) and Lung Cancer (C), the agents infer edges forming between S and C, and G and C, because they are not conditionally independent. No edge forms between S and G as they are assumed to be independent. Once the skeletons are identified, the second principle aims to discover the directions they take. Specifically, if there is no edge between A and C but there are edges between A and B and B and C, then B is a collider (A $\rightarrow$ B $\leftarrow$ C) if and only if A and C, though marginally independent, become dependent when conditioned on B. Following the previous example, S and G are independent, but if we know someone has Lung Cancer (C), learning about their smoking status (S) gives information about their likely genetic predisposition (G), and vice-versa. This means C is a collider, so we have S → C ← G. In other situations, the causal direction cannot be identified by conditional independence alone; it can only be determined by the acyclic constraint and the causal order derived from the combination of metadata (Principle 3). Back to previous example, the collider rule directed all edges and the structure S → C ← G has no cycles, therefore, this is a valid DAG. 

The \textbf{Procedure} section outlines essential guidelines for agents to apply the causal principles in a debating format. Specifically, after Causal Affirmative Agent proposes the first graph, Causal Negative Agent is prompted to propose different causal networks while adhering to the principles and incorporating metadata as much as possible. The differences are primarily reflected in three aspects: the skeleton, colliders, and the direction of other edges. This results in significant differences between the agents representing the affirmative and negative sides of the debate. Lastly, Causal Judge uses a quantifiable metric to determine which causal graph presented by the affirmative and negative debaters is more reasonable. It also considers all the differences between the causal graphs proposed by both sides in terms of their skeleton, colliders, and the direction of other causal edges.  The overall comparison will reveal which graph aligns more closely with the metadata, leading to the final conclusion.

Finally, the \textbf{Output Format} ensures that the agents output in a desired format for parsing the results.

\subsection{Design of MDM with SCD Algorithm as Output}\label{A-prompt-MDM-SCD}

\begin{figure*}[t]
    \centering
    \includegraphics[width=\linewidth]{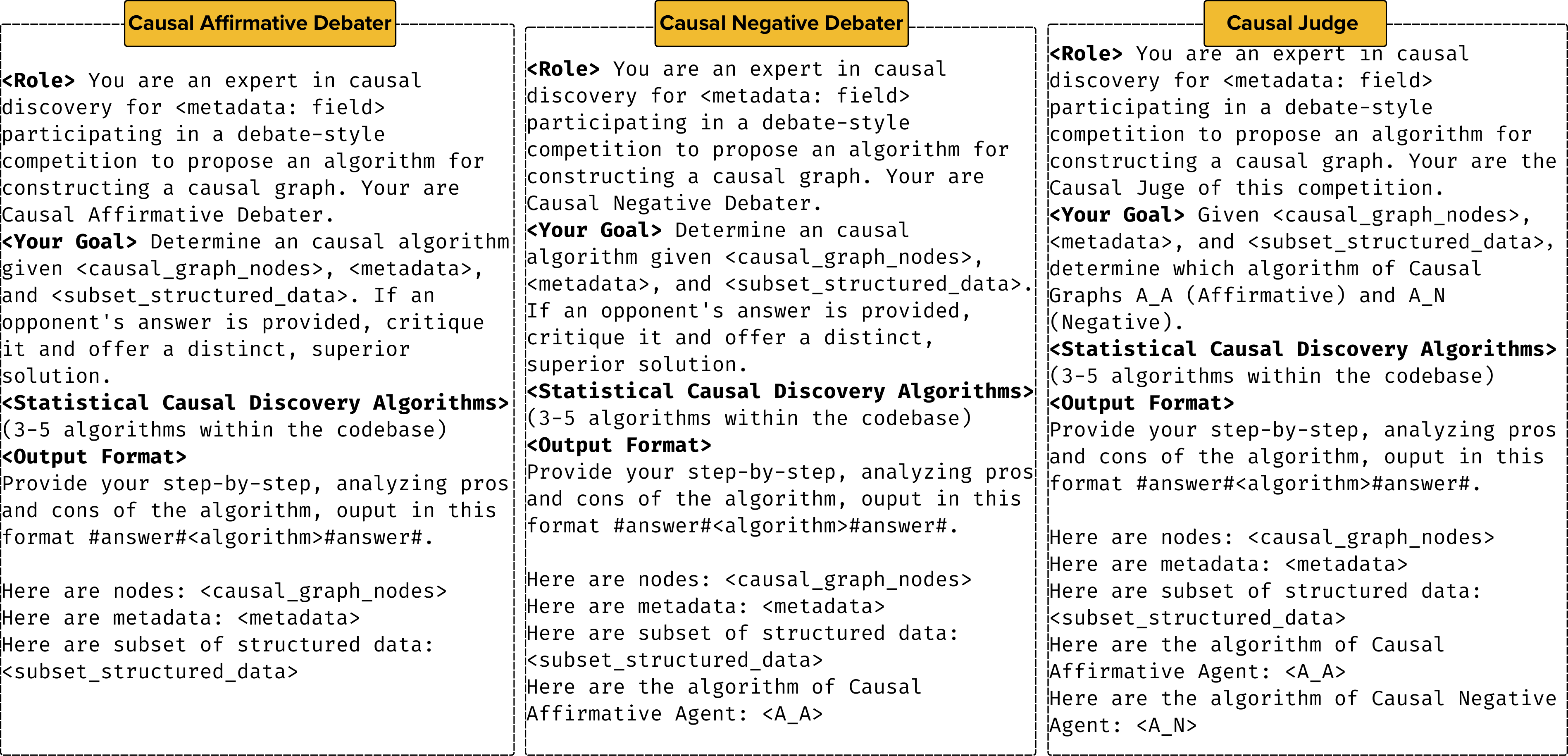}
    \caption{Prompt templates for the three MDM agents when MDM outputs an SCD algorithm (\texttt{alg} mode), used inside DCM for algorithm selection. In addition to the graph nodes $V$ and metadata $I$, each agent receives a small subset of structured data $S$ and a pool of candidate SCD algorithms, then argues for and adjudicates the most suitable choice.}
    \label{fig:prompt-mdm-alg}
\end{figure*}

Similar to the prompt design for MDM with causal graph as output, the MDM framework with  Statistical Causal Discovery (SCD) algorithm as output also begins by explicitly identifying the \textbf{Role} of each agent (full templates in Figure~\ref{fig:prompt-mdm-alg}). However, it accepts additional inputs—such as \(S\)—and pursues a different goal. Specifically, in the \textbf{Your Goal} section, each agent is asked to select the most suitable statistical causal discovery algorithm based on dataset information (\(V\), \(I\), \(S\)).

Each agent is pre-prompted with several \textbf{Statistical Causal Discovery Algorithms}, such as the PC algorithm, FCI, and GES. Based on the dataset characteristics (\(V\), \(I\), \(S\)), agents propose different algorithms. For instance, Causal Affirmative Debater might argue that for 20 continuous temperature variables (\(V\), \(I\), \(S\)), GES is the best choice because it is well-suited for continuous data, can handle 20 variables, and aims to provide a clear, affirmative causal model. In contrast, Causal Negative Debater may counter that while GES is an option, temperature data (\(S\)) often includes unmeasured confounders, which GES does not address. To be more cautious with these 20 variables (\(V\)), Causal Negative Debater suggests FCI, which is designed to be robust against hidden confounders and therefore yields more reliable, though potentially less specific, causal insights. Causal Judge then evaluates these perspectives. Causal Affirmative Debater correctly points out GES's strengths for the specified continuous 20-variable data (\(V\), \(I\), \(S\)), while Causal Negative Debater brings up a valid concern about potential hidden confounders within the temperature data subset (\(S\)), justifying a preference for FCI's robustness. Causal Judge's decision ultimately depends on the metadata (\(I\)) associated with subset \(S\). If \(I\) provides strong evidence that there are no significant unmeasured confounders among the 20 variables (\(V\)) in \(S\), then GES is chosen. Conversely, if the metadata (\(I\)) is ambiguous regarding confounders or suggests they might be present in \(S\), FCI is selected to prioritize safety and reliability. In the absence of strong evidence from \(I\) ruling out confounders for this specific data subset (\(S\)), Causal Judge defaults to selecting FCI.

Finally, the \textbf{Output Format} also ensures that the agents output in a desired format for parsing the results.

\section{Evaluation Metrics}\label{A-Evaluation Metrics}

We assess the final learned adjacency matrix of the causal graph by comparing it with the true adjacency matrix using the following metrics: Structural Hamming Distance (SHD) \cite{takayama2024integrating}, Normalized Hamming Distance (NHD) \cite{kıcıman2023causal}, and F1-Score.

\subsection{Structural Hamming Distance (SHD)}
The Structural Hamming Distance (SHD) for two directed graphs \( G \) and \( G' \) with \( m \) nodes is given by:
\[
\text{SHD}(G, G') = \sum_{1 \leq i \neq j \leq m} \mathbb{1}_{G_{ij} \neq G'_{ij}}
\]
where \( \mathbb{1}_{G_{ij} \neq G'_{ij}} \) is an indicator function that equals 1 if the directed edge from node \( i \) to node \( j \) is present in one graph but not the other, and 0 otherwise.

This metric counts the total number of directed edge disagreements between the two graphs, providing a measure of structural difference without normalization.

\subsection{Normalized Hamming Distance (NHD)}
The Normalized Hamming Distance (NHD) for two undirected graphs \( G \) and \( G' \) with \( m \) nodes is given by:
\[
\text{NHD}(G, G') = \frac{1}{\binom{m}{2}} \sum_{1 \leq i < j \leq m} \mathbb{1}_{G_{ij} \neq G'_{ij}}
\]
where \( \binom{m}{2} = \frac{m(m-1)}{2} \) is the total number of possible edges in an undirected graph with \( m \) nodes. \( \mathbb{1}_{G_{ij} \neq G'_{ij}} \) is an indicator function that equals $1$ if the edge between nodes \( i \) and \( j \) exist in one graph but not the other, and $0$ otherwise.

This metric measures the proportion of differing edges relative to the total number of possible edges, providing a value between 0 and 1, where 0 indicates identical graphs and 1 indicates completely dissimilar graphs.

\subsection{F1 Score}
The F1 Score for comparing the edges of two graphs \( G \) and \( G' \) is given by:
\[
\text{F1}(G, G') = 2 \cdot \frac{\text{precision} \cdot \text{recall}}{\text{precision} + \text{recall}}
\]
where:
\vspace{-0.3cm}
\begin{itemize}
    \item For directed graphs:
    \begin{itemize}
\vspace{-0.3cm}
    
        \item True Positives (TP): \( \sum_{i \neq j} \mathbb{1}_{G_{ij} = 1 \text{ and } G'_{ij} = 1} \)
        \item False Positives (FP): \( \sum_{i \neq j} \mathbb{1}_{G_{ij} = 0 \text{ and } G'_{ij} = 1} \)
        \item False Negatives (FN): \( \sum_{i \neq j} \mathbb{1}_{G_{ij} = 1 \text{ and } G'_{ij} = 0} \)
    \end{itemize}
    \item Precision: \( \frac{\text{TP}}{\text{TP} + \text{FP}} \)
    \item Recall: \( \frac{\text{TP}}{\text{TP} + \text{FN}} \)
\end{itemize}
The F1 Score provides a balanced measure of precision and recall, adapted to the context of graph edge comparison.

\section{More Related Work}

\subsection{Prompt Engineering}Several advanced techniques for leveraging large language models (LLMs) have been identified. Zero-shot prompting, introduced by \cite{Radford2019LanguageMA}, guides LLMs to perform novel tasks using carefully crafted prompts without the need for training data, allowing the model to leverage its existing knowledge to generate predictions. Few-shot prompting, as described by \cite{brown2020language}, enhances model performance by providing a few input-output examples, though it requires more tokens and careful example selection to mitigate biases. For reasoning and logic, \cite{wei2023chainofthought} introduced Chain-of-Thought (CoT) prompting, which guides LLMs through step-by-step reasoning processes, significantly improving accuracy in complex tasks such as math and commonsense reasoning. Building on this, \cite{wang2023selfconsistency} proposed self-consistency, a strategy that generates diverse reasoning chains and identifies the most consistent final answer, further enhancing accuracy. Additionally, \cite{yao2023react} developed ReAct, enabling LLMs to generate reasoning traces and task-specific actions concurrently, thereby improving performance in question answering, fact verification, and interactive decision-making by enhancing the synergy between reasoning and action. In this work, we design the prompt to adhere to key principles, ensuring the correct temporal order of variables and maintaining an acyclic structure.

\subsection{LLMs' Agentic Workflow}
A general LLM agent framework consists of core components: user request, agent/brain, planning, memory, and tools. The agent/brain acts as the main coordinator, activated by a prompt template. It can be profiled with specific details to define its role, using handcrafted, LLM-generated, or data-driven strategies. Planning employs techniques like Chain of Thought and Tree of Thoughts, and for complex tasks, feedback mechanisms like ReAct \cite{yao2023react} and Reflexion \cite{shinn2023reflexion} refine plans based on past actions and observations. Memory stores the agent's logs, with short-term memory for the current context and long-term memory for past behaviors. Hybrid memory combines both to enhance reasoning and experience accumulation. Tools enable interaction with external environments, such as APIs and code interpreters. Frameworks like MRKL \cite{karpas2022mrkl}, Toolformer \cite{schick2023toolformer}, Function Calling \cite{openai_function_calling}, and HuggingGPT \cite{shen2023hugginggpt} integrate tools to solve tasks effectively.

However, for more complex problems where a single LLM agent may struggle, LLM-based multi-agent (LLM-MA) systems excel \citep{guo2024multiagentsurvey, wang2024llmagentsurvey}. Current LLM-MA systems primarily employ three communication paradigms: Cooperative, Competitive, and Debating. In the Cooperative paradigm, agents collaborate towards a shared goal, typically exchanging information to enhance a collective solution; role-specialized frameworks such as MetaGPT \citep{hong2024metagpt} and AgentVerse \citep{chen2024agentverse} show that structured cooperation reduces cascading errors and outperforms monolithic models \citep{qian2023communicative, chen2024scalable}. In the Competitive paradigm, agents work towards their own goals, which might conflict with those of other agents \citep{zhao2023competeai}. The Debating paradigm involves agents engaging in argumentative interactions, where they present and defend their viewpoints while critiquing those of others; this approach is effective for reaching a consensus or a more refined solution and has been shown to mitigate hallucination and improve collective reasoning \citep{du2023improving, zhang2024machinesom, chan2024chateval, li2023camel, liang2023encouraging, xiong2023examining}. In this work, the debating paradigm is adopted, as the nature of causal discovery requires diverse and potentially conflicting hypotheses to be adjudicated before converging on the truth.

\section{Dataset}\label{A-datasets}
We measure the performance of our MAC framework using five different datasets. The details of each dataset are as follows:
\vspace{-0.3cm}
\begin{itemize}
    \item \textbf{Auto} \cite{misc_auto_mpg_9}: It is from the UCI Machine Learning Repository and is a commonly used benchmark for causal inference. It includes variables related to car fuel consumption with five variables selected, “Weight,” “Displacement,” “Horsepower,” “Acceleration,” and “MPG” (miles per gallon) for causal inference. It has 392 continuous observations.
\vspace{-0.3cm}
    \item \textbf{Child} \cite{spiegelhalter1992learning}: It is a moderately sized dataset that focuses on congenital heart disease in newborns. The ground-truth graph consists of 20 nodes with 25 edges. Some key variables in this graph include Birth Asphyxia, Lung Flow, and Chest X-Ray. It has 10,000 integer observations.
 \vspace{-0.3cm}
    \item \textbf{Cancer} \cite{korb2010bayesian}: It is a medical dataset, in the context of metastatic cancer. The ground-truth graph consists of five nodes, Pollution, Cancer, Smoker, X-ray, and Dyspnoea, with four edges. It has 10,000 binary observations. 
\vspace{-0.3cm}
    \item \textbf{Earthquake} \cite{korb2010bayesian}: It is a dataset related to earthquake detection. The ground-truth graph consists of five nodes, Burglary, Alarm, Earthquake, John Calls, and Mary Calls, with four edges. It includes 10,000 binary observations.
\vspace{-0.3cm}
    \item \textbf{Survey} \cite{scutari2021bayesian}: It is a dataset related to demographic and social behaviors. The ground-truth graph consists of six nodes, Age, Education, Sex, Occupation, Residence, and Travel, with six edges. It includes 10,000 integer observations.
\end{itemize}

\section{Additional Experiment Results}\label{A-Additional Results}

This section comprises the detailed experiment results for all of the models. We note that the DCM rows are identical across the three backbones: for each dataset, all three LLMs selected the same SCD algorithm during the debate, and since the selected algorithm is then executed deterministically on the full data, the resulting DCM graphs---and hence their metrics---coincide.

\begin{table*}[t]
\centering
\caption{Overall results across five datasets and three backbone LLMs (mean $\pm$ std over 10 seeds). For each metric within a model, \textbf{bold} marks the best method and \underline{underline} the second best. Lower SHD/NHD and higher F1 are better. The five statistical baselines (Exact Search, GES, PC, FCI, LiNGAM) do not use an LLM, so their values are identical across the three models.}
\label{tab:overall-merged}
\resizebox{\textwidth}{!}{%
\begin{tabular}{ll|ccc|ccc|ccc}
\toprule
\multirow{2}{*}{\textbf{Dataset}} & \multirow{2}{*}{\textbf{Method}} & \multicolumn{3}{c|}{\textbf{Gemini-2.0-Flash}} & \multicolumn{3}{c|}{\textbf{DeepSeek-R1}} & \multicolumn{3}{c}{\textbf{GPT-4o}} \\
 & & SHD$\downarrow$ & NHD$\downarrow$ & F1$\uparrow$ & SHD$\downarrow$ & NHD$\downarrow$ & F1$\uparrow$ & SHD$\downarrow$ & NHD$\downarrow$ & F1$\uparrow$ \\
\midrule
\multirow{10}{*}{Child} & Exact Search & \underline{25.00} $\pm$ 0.00 & \textbf{0.06} $\pm$ 0.00 & 0.00 $\pm$ 0.00 & 25.00 $\pm$ 0.00 & \underline{0.06} $\pm$ 0.00 & 0.00 $\pm$ 0.00 & 25.00 $\pm$ 0.00 & \underline{0.062} $\pm$ 0.000 & 0.00 $\pm$ 0.00 \\
 & GES & 39.33 $\pm$ 1.70 & 0.15 $\pm$ 0.00 & 0.20 $\pm$ 0.05 & 39.33 $\pm$ 1.70 & 0.15 $\pm$ 0.00 & 0.20 $\pm$ 0.05 & 39.33 $\pm$ 1.70 & 0.150 $\pm$ 0.000 & 0.20 $\pm$ 0.05 \\
 & PC & 30.67 $\pm$ 2.05 & 0.12 $\pm$ 0.00 & 0.36 $\pm$ 0.03 & 30.67 $\pm$ 2.05 & 0.12 $\pm$ 0.00 & 0.36 $\pm$ 0.03 & 30.67 $\pm$ 2.05 & 0.120 $\pm$ 0.000 & 0.36 $\pm$ 0.03 \\
 & FCI & 39.67 $\pm$ 0.94 & 0.17 $\pm$ 0.00 & 0.33 $\pm$ 0.02 & 39.67 $\pm$ 0.94 & 0.17 $\pm$ 0.00 & 0.33 $\pm$ 0.02 & 39.67 $\pm$ 0.94 & 0.170 $\pm$ 0.000 & 0.33 $\pm$ 0.02 \\
 & LiNGAM & 64.33 $\pm$ 2.62 & 0.19 $\pm$ 0.01 & 0.26 $\pm$ 0.01 & 64.33 $\pm$ 2.62 & 0.19 $\pm$ 0.01 & 0.26 $\pm$ 0.01 & 64.33 $\pm$ 2.62 & 0.190 $\pm$ 0.010 & 0.26 $\pm$ 0.01 \\
 & Zero-shot & 27.00 $\pm$ 0.00 & 0.08 $\pm$ 0.00 & 0.33 $\pm$ 0.00 & 21.33 $\pm$ 1.70 & \underline{0.06} $\pm$ 0.01 & 0.57 $\pm$ 0.06 & 28.67 $\pm$ 2.36 & 0.080 $\pm$ 0.010 & 0.35 $\pm$ 0.07 \\
 & ReAct & 27.67 $\pm$ 2.36 & \textbf{0.06} $\pm$ 0.01 & 0.40 $\pm$ 0.05 & 21.67 $\pm$ 5.73 & \underline{0.06} $\pm$ 0.01 & 0.55 $\pm$ 0.12 & 23.00 $\pm$ 2.94 & \textbf{0.061} $\pm$ 0.010 & 0.35 $\pm$ 0.08 \\
 & LLM-SCD & 25.15 $\pm$ 2.50 & 0.08 $\pm$ 0.01 & \underline{0.41} $\pm$ 0.08 & \underline{16.82} $\pm$ 3.15 & \underline{0.06} $\pm$ 0.02 & \underline{0.61} $\pm$ 0.08 & \underline{21.15} $\pm$ 1.30 & 0.070 $\pm$ 0.020 & \underline{0.39} $\pm$ 0.03 \\
 & ILS-CSL & 26.50 $\pm$ 3.00 & 0.09 $\pm$ 0.02 & 0.39 $\pm$ 0.10 & 17.54 $\pm$ 3.50 & 0.07 $\pm$ 0.03 & 0.59 $\pm$ 0.10 & 22.40 $\pm$ 1.45 & 0.080 $\pm$ 0.020 & 0.37 $\pm$ 0.04 \\
\cmidrule(l){2-11}
 & \textbf{MAC} & \textbf{24.33} $\pm$ 3.09 & \underline{0.07} $\pm$ 0.01 & \textbf{0.44} $\pm$ 0.09 & \textbf{15.67} $\pm$ 3.09 & \textbf{0.05} $\pm$ 0.01 & \textbf{0.65} $\pm$ 0.07 & \textbf{20.33} $\pm$ 1.25 & \textbf{0.061} $\pm$ 0.010 & \textbf{0.41} $\pm$ 0.02 \\
\midrule
\multirow{10}{*}{Auto} & Exact Search & 7.00 $\pm$ 0.00 & 0.44 $\pm$ 0.00 & 0.15 $\pm$ 0.00 & 7.00 $\pm$ 0.00 & 0.44 $\pm$ 0.00 & 0.15 $\pm$ 0.00 & 7.00 $\pm$ 0.00 & 0.44 $\pm$ 0.00 & 0.15 $\pm$ 0.00 \\
 & GES & \textbf{3.00} $\pm$ 0.00 & 0.32 $\pm$ 0.00 & 0.57 $\pm$ 0.00 & \textbf{3.00} $\pm$ 0.00 & 0.32 $\pm$ 0.00 & \textbf{0.57} $\pm$ 0.00 & \textbf{3.00} $\pm$ 0.00 & 0.32 $\pm$ 0.00 & \textbf{0.57} $\pm$ 0.00 \\
 & PC & 8.00 $\pm$ 0.00 & 0.48 $\pm$ 0.00 & 0.14 $\pm$ 0.00 & 8.00 $\pm$ 0.00 & 0.48 $\pm$ 0.00 & 0.14 $\pm$ 0.00 & 8.00 $\pm$ 0.00 & 0.48 $\pm$ 0.00 & 0.14 $\pm$ 0.00 \\
 & FCI & 5.00 $\pm$ 0.00 & 0.24 $\pm$ 0.00 & 0.25 $\pm$ 0.00 & 5.00 $\pm$ 0.00 & 0.24 $\pm$ 0.00 & 0.25 $\pm$ 0.00 & \underline{5.00} $\pm$ 0.00 & \underline{0.24} $\pm$ 0.00 & 0.25 $\pm$ 0.00 \\
 & LiNGAM & 8.00 $\pm$ 0.00 & 0.48 $\pm$ 0.00 & 0.14 $\pm$ 0.00 & 8.00 $\pm$ 0.00 & 0.48 $\pm$ 0.00 & 0.14 $\pm$ 0.00 & 8.00 $\pm$ 0.00 & 0.48 $\pm$ 0.00 & 0.14 $\pm$ 0.00 \\
 & Zero-shot & 7.00 $\pm$ 0.00 & 0.32 $\pm$ 0.00 & 0.20 $\pm$ 0.00 & 7.00 $\pm$ 0.82 & 0.33 $\pm$ 0.05 & 0.29 $\pm$ 0.09 & 6.75 $\pm$ 1.09 & 0.32 $\pm$ 0.06 & 0.26 $\pm$ 0.08 \\
 & ReAct & 7.67 $\pm$ 0.47 & 0.35 $\pm$ 0.02 & 0.32 $\pm$ 0.01 & 6.67 $\pm$ 0.47 & 0.32 $\pm$ 0.03 & 0.33 $\pm$ 0.02 & 6.33 $\pm$ 0.47 & 0.28 $\pm$ 0.03 & 0.37 $\pm$ 0.03 \\
 & LLM-SCD & 5.10 $\pm$ 0.50 & \underline{0.22} $\pm$ 0.05 & \underline{0.58} $\pm$ 0.07 & 4.95 $\pm$ 1.10 & \underline{0.23} $\pm$ 0.05 & 0.53 $\pm$ 0.09 & 5.55 $\pm$ 1.70 & 0.25 $\pm$ 0.06 & 0.41 $\pm$ 0.19 \\
 & ILS-CSL & 5.50 $\pm$ 0.60 & 0.25 $\pm$ 0.06 & 0.55 $\pm$ 0.09 & 5.41 $\pm$ 1.25 & 0.25 $\pm$ 0.06 & 0.51 $\pm$ 0.11 & 5.95 $\pm$ 1.85 & 0.27 $\pm$ 0.07 & 0.39 $\pm$ 0.21 \\
\cmidrule(l){2-11}
 & \textbf{MAC} & \underline{4.33} $\pm$ 0.47 & \textbf{0.19} $\pm$ 0.04 & \textbf{0.61} $\pm$ 0.08 & \underline{4.33} $\pm$ 0.94 & \textbf{0.21} $\pm$ 0.04 & \underline{0.56} $\pm$ 0.08 & \underline{5.00} $\pm$ 1.63 & \textbf{0.23} $\pm$ 0.05 & \underline{0.43} $\pm$ 0.18 \\
\midrule
\multirow{10}{*}{Earthquake} & Exact Search & \textbf{0.00} $\pm$ 0.00 & \textbf{0.00} $\pm$ 0.00 & \textbf{1.00} $\pm$ 0.00 & \textbf{0.00} $\pm$ 0.00 & \textbf{0.00} $\pm$ 0.00 & \textbf{1.00} $\pm$ 0.00 & \textbf{0.00} $\pm$ 0.00 & \textbf{0.00} $\pm$ 0.00 & \textbf{1.00} $\pm$ 0.00 \\
 & GES & 4.00 $\pm$ 0.00 & 0.32 $\pm$ 0.00 & 0.00 $\pm$ 0.00 & 4.00 $\pm$ 0.00 & 0.32 $\pm$ 0.00 & 0.00 $\pm$ 0.00 & 4.00 $\pm$ 0.00 & 0.32 $\pm$ 0.00 & 0.00 $\pm$ 0.00 \\
 & PC & 4.67 $\pm$ 0.00 & 0.35 $\pm$ 0.00 & 0.00 $\pm$ 0.00 & 4.67 $\pm$ 0.00 & 0.35 $\pm$ 0.00 & 0.00 $\pm$ 0.00 & 4.67 $\pm$ 0.00 & 0.35 $\pm$ 0.00 & 0.00 $\pm$ 0.00 \\
 & FCI & 4.33 $\pm$ 0.00 & 0.25 $\pm$ 0.00 & 0.00 $\pm$ 0.00 & 4.33 $\pm$ 0.00 & 0.25 $\pm$ 0.00 & 0.00 $\pm$ 0.00 & 4.33 $\pm$ 0.00 & 0.25 $\pm$ 0.00 & 0.00 $\pm$ 0.00 \\
 & LiNGAM & 4.00 $\pm$ 0.00 & 0.28 $\pm$ 0.00 & 0.22 $\pm$ 0.00 & 4.00 $\pm$ 0.00 & 0.28 $\pm$ 0.00 & 0.22 $\pm$ 0.00 & 4.00 $\pm$ 0.00 & 0.28 $\pm$ 0.00 & 0.22 $\pm$ 0.00 \\
 & Zero-shot & 1.00 $\pm$ 0.00 & 0.04 $\pm$ 0.00 & 0.89 $\pm$ 0.00 & 1.00 $\pm$ 0.00 & 0.04 $\pm$ 0.00 & 0.89 $\pm$ 0.00 & 1.00 $\pm$ 0.00 & 0.04 $\pm$ 0.00 & 0.89 $\pm$ 0.00 \\
 & ReAct & 1.00 $\pm$ 0.00 & 0.04 $\pm$ 0.00 & 0.89 $\pm$ 0.00 & 1.00 $\pm$ 0.00 & 0.04 $\pm$ 0.00 & 0.89 $\pm$ 0.00 & 1.00 $\pm$ 0.00 & 0.04 $\pm$ 0.00 & 0.89 $\pm$ 0.00 \\
 & LLM-SCD & \underline{0.50} $\pm$ 0.10 & \underline{0.01} $\pm$ 0.01 & \underline{0.95} $\pm$ 0.02 & \underline{0.45} $\pm$ 0.15 & \underline{0.01} $\pm$ 0.01 & \underline{0.96} $\pm$ 0.03 & 0.40 $\pm$ 0.50 & 0.02 $\pm$ 0.02 & 0.94 $\pm$ 0.06 \\
 & ILS-CSL & 0.80 $\pm$ 0.20 & 0.02 $\pm$ 0.01 & 0.92 $\pm$ 0.03 & 0.78 $\pm$ 0.25 & 0.02 $\pm$ 0.01 & 0.93 $\pm$ 0.04 & 0.50 $\pm$ 0.60 & 0.03 $\pm$ 0.03 & 0.92 $\pm$ 0.07 \\
\cmidrule(l){2-11}
 & \textbf{MAC} & \textbf{0.00} $\pm$ 0.00 & \textbf{0.00} $\pm$ 0.00 & \textbf{1.00} $\pm$ 0.00 & \textbf{0.00} $\pm$ 0.00 & \textbf{0.00} $\pm$ 0.00 & \textbf{1.00} $\pm$ 0.00 & \underline{0.33} $\pm$ 0.47 & \underline{0.01} $\pm$ 0.02 & \underline{0.96} $\pm$ 0.05 \\
\midrule
\multirow{10}{*}{Cancer} & Exact Search & 2.33 $\pm$ 0.00 & 0.17 $\pm$ 0.00 & 0.44 $\pm$ 0.00 & 2.33 $\pm$ 0.00 & 0.17 $\pm$ 0.00 & 0.44 $\pm$ 0.00 & 2.33 $\pm$ 0.00 & 0.17 $\pm$ 0.00 & 0.44 $\pm$ 0.00 \\
 & GES & \textbf{1.33} $\pm$ 0.00 & 0.17 $\pm$ 0.00 & 0.62 $\pm$ 0.00 & 1.33 $\pm$ 0.00 & 0.17 $\pm$ 0.00 & 0.62 $\pm$ 0.00 & \underline{1.33} $\pm$ 0.00 & 0.17 $\pm$ 0.00 & 0.62 $\pm$ 0.00 \\
 & PC & \underline{2.00} $\pm$ 0.00 & 0.16 $\pm$ 0.00 & 0.50 $\pm$ 0.00 & 2.00 $\pm$ 0.00 & 0.16 $\pm$ 0.00 & 0.50 $\pm$ 0.00 & 2.00 $\pm$ 0.00 & 0.16 $\pm$ 0.00 & 0.50 $\pm$ 0.00 \\
 & FCI & 4.00 $\pm$ 0.00 & 0.16 $\pm$ 0.00 & 0.00 $\pm$ 0.00 & 4.00 $\pm$ 0.00 & 0.16 $\pm$ 0.00 & 0.00 $\pm$ 0.00 & 4.00 $\pm$ 0.00 & 0.16 $\pm$ 0.00 & 0.00 $\pm$ 0.00 \\
 & LiNGAM & 2.33 $\pm$ 0.00 & 0.17 $\pm$ 0.00 & 0.43 $\pm$ 0.00 & 2.33 $\pm$ 0.00 & 0.17 $\pm$ 0.00 & 0.43 $\pm$ 0.00 & 2.33 $\pm$ 0.00 & 0.17 $\pm$ 0.00 & 0.43 $\pm$ 0.00 \\
 & Zero-shot & 3.00 $\pm$ 0.00 & 0.12 $\pm$ 0.00 & 0.73 $\pm$ 0.00 & 1.33 $\pm$ 0.47 & 0.05 $\pm$ 0.02 & 0.86 $\pm$ 0.04 & 2.00 $\pm$ 0.00 & 0.08 $\pm$ 0.00 & 0.80 $\pm$ 0.00 \\
 & ReAct & \underline{2.00} $\pm$ 0.00 & \textbf{0.08} $\pm$ 0.00 & \textbf{0.80} $\pm$ 0.00 & \textbf{0.67} $\pm$ 0.47 & \textbf{0.03} $\pm$ 0.02 & \textbf{0.93} $\pm$ 0.05 & \textbf{1.00} $\pm$ 0.00 & \textbf{0.04} $\pm$ 0.00 & \textbf{0.89} $\pm$ 0.00 \\
 & LLM-SCD & 2.50 $\pm$ 0.10 & \underline{0.10} $\pm$ 0.01 & \underline{0.75} $\pm$ 0.02 & 1.45 $\pm$ 0.90 & 0.06 $\pm$ 0.04 & 0.87 $\pm$ 0.09 & 1.85 $\pm$ 1.75 & 0.08 $\pm$ 0.08 & 0.81 $\pm$ 0.18 \\
 & ILS-CSL & 2.80 $\pm$ 0.20 & 0.12 $\pm$ 0.02 & 0.72 $\pm$ 0.03 & 1.83 $\pm$ 1.10 & 0.07 $\pm$ 0.05 & 0.84 $\pm$ 0.11 & 2.10 $\pm$ 1.90 & 0.09 $\pm$ 0.09 & 0.79 $\pm$ 0.20 \\
\cmidrule(l){2-11}
 & \textbf{MAC} & \underline{2.00} $\pm$ 0.00 & \textbf{0.08} $\pm$ 0.00 & \textbf{0.80} $\pm$ 0.00 & \underline{1.00} $\pm$ 0.82 & \underline{0.04} $\pm$ 0.03 & \underline{0.90} $\pm$ 0.08 & 1.67 $\pm$ 1.70 & \underline{0.07} $\pm$ 0.07 & \underline{0.83} $\pm$ 0.17 \\
\midrule
\multirow{10}{*}{Survey} & Exact Search & 4.33 $\pm$ 0.00 & 0.21 $\pm$ 0.00 & 0.28 $\pm$ 0.00 & 4.33 $\pm$ 0.00 & 0.21 $\pm$ 0.00 & 0.28 $\pm$ 0.00 & 4.33 $\pm$ 0.00 & 0.21 $\pm$ 0.00 & 0.28 $\pm$ 0.00 \\
 & GES & \underline{3.00} $\pm$ 0.00 & 0.17 $\pm$ 0.00 & 0.54 $\pm$ 0.00 & \underline{3.00} $\pm$ 0.00 & 0.17 $\pm$ 0.00 & 0.54 $\pm$ 0.00 & 3.00 $\pm$ 0.00 & 0.17 $\pm$ 0.00 & 0.54 $\pm$ 0.00 \\
 & PC & 4.33 $\pm$ 0.00 & 0.22 $\pm$ 0.00 & 0.42 $\pm$ 0.00 & 4.33 $\pm$ 0.00 & 0.22 $\pm$ 0.00 & 0.42 $\pm$ 0.00 & 4.33 $\pm$ 0.00 & 0.22 $\pm$ 0.00 & 0.42 $\pm$ 0.00 \\
 & FCI & 5.00 $\pm$ 0.00 & 0.19 $\pm$ 0.00 & 0.22 $\pm$ 0.00 & 5.00 $\pm$ 0.00 & 0.19 $\pm$ 0.00 & 0.22 $\pm$ 0.00 & 5.00 $\pm$ 0.00 & 0.19 $\pm$ 0.00 & 0.22 $\pm$ 0.00 \\
 & LiNGAM & \textbf{2.67} $\pm$ 0.00 & \textbf{0.11} $\pm$ 0.00 & 0.65 $\pm$ 0.00 & \textbf{2.67} $\pm$ 0.00 & \textbf{0.11} $\pm$ 0.00 & 0.65 $\pm$ 0.00 & 2.67 $\pm$ 0.00 & 0.11 $\pm$ 0.00 & 0.65 $\pm$ 0.00 \\
 & Zero-shot & 4.00 $\pm$ 0.00 & \textbf{0.11} $\pm$ 0.00 & \underline{0.67} $\pm$ 0.00 & 7.00 $\pm$ 1.63 & 0.19 $\pm$ 0.05 & 0.55 $\pm$ 0.13 & 7.00 $\pm$ 0.82 & 0.19 $\pm$ 0.02 & 0.58 $\pm$ 0.10 \\
 & ReAct & 6.00 $\pm$ 0.00 & 0.17 $\pm$ 0.00 & 0.60 $\pm$ 0.04 & 8.00 $\pm$ 0.82 & 0.22 $\pm$ 0.02 & 0.45 $\pm$ 0.06 & 7.33 $\pm$ 1.25 & 0.20 $\pm$ 0.03 & 0.45 $\pm$ 0.04 \\
 & LLM-SCD & 4.50 $\pm$ 0.80 & 0.16 $\pm$ 0.01 & 0.66 $\pm$ 0.02 & 4.67 $\pm$ 1.55 & \underline{0.13} $\pm$ 0.05 & \underline{0.68} $\pm$ 0.11 & \underline{2.65} $\pm$ 0.50 & \underline{0.07} $\pm$ 0.02 & \underline{0.77} $\pm$ 0.02 \\
 & ILS-CSL & 4.90 $\pm$ 0.90 & 0.18 $\pm$ 0.02 & 0.63 $\pm$ 0.03 & 5.21 $\pm$ 1.80 & 0.15 $\pm$ 0.06 & 0.65 $\pm$ 0.13 & 2.90 $\pm$ 0.60 & 0.08 $\pm$ 0.02 & 0.75 $\pm$ 0.03 \\
\cmidrule(l){2-11}
 & \textbf{MAC} & 4.00 $\pm$ 0.82 & \underline{0.14} $\pm$ 0.00 & \textbf{0.69} $\pm$ 0.02 & 4.00 $\pm$ 1.41 & \textbf{0.11} $\pm$ 0.04 & \textbf{0.71} $\pm$ 0.10 & \textbf{2.33} $\pm$ 0.47 & \textbf{0.06} $\pm$ 0.01 & \textbf{0.79} $\pm$ 0.01 \\
\bottomrule
\end{tabular}}
\end{table*}


\begin{table*}[]
\centering 
\caption{Comparison of methods on Child, Auto, Earthquake, Cancer, and Survey (with standard deviations) with other modules of MAC framework (Gemini Flash 2.0).}
\label{gemini-combined-restructured} 
\begin{tabular}{llccc} 
\toprule
\textbf{Dataset} & \textbf{Method} & \textbf{SHD $(\downarrow)$} & \textbf{NHD $(\downarrow)$} & \textbf{F1 $(\uparrow)$} \\
\midrule
 & DCM & 30.67 $\pm$ 11.09 & 0.15 $\pm$ 0.00 & 0.19 $\pm$ 0.06 \\
Child & MDM & 30.67 $\pm$ 0.47 & 0.08 $\pm$ 0.00 & 0.33 $\pm$ 0.00 \\
 & MAC & \textbf{24.33} $\pm$ 3.09 & \textbf{0.07} $\pm$ 0.01 & \textbf{0.44} $\pm$ 0.09 \\
\midrule 
 & DCM & \textbf{4.67} $\pm$ 1.25 & 0.35 $\pm$ 0.10 & 0.31 $\pm$ 0.14 \\
Auto & MDM & 7.67 $\pm$ 0.47 & 0.35 $\pm$ 0.02 & 0.32 $\pm$ 0.01 \\
 & MAC & 4.33 $\pm$ 0.47 & \textbf{0.19} $\pm$ 0.04 & \textbf{0.61} $\pm$ 0.08 \\
\midrule
 & DCM & 4.67 $\pm$ 0.47 & 0.32 $\pm$ 0.03 & 0.00 $\pm$ 0.00 \\
Earthquake & MDM & 0.67 $\pm$ 0.47 & 0.03 $\pm$ 0.02 & 0.93 $\pm$ 0.05 \\
 & MAC & \textbf{0.00} $\pm$ 0.00 & \textbf{0.00} $\pm$ 0.00 & \textbf{1.00} $\pm$ 0.00 \\
\midrule
 & DCM & 3.00 $\pm$ 0.82 & 0.17 $\pm$ 0.02 & 0.26 $\pm$ 0.20 \\
Cancer & MDM & \textbf{1.00} $\pm$ 0.00 & \textbf{0.04} $\pm$ 0.00 & \textbf{0.89} $\pm$ 0.00 \\
 & MAC & 2.00 $\pm$ 0.00 & 0.08 $\pm$ 0.00 & 0.80 $\pm$ 0.00 \\
\midrule
 & DCM & \textbf{2.00} $\pm$ 0.00 & 0.22 $\pm$ 0.08 & 0.25 $\pm$ 0.35 \\
Survey & MDM & 6.00 $\pm$ 1.41 & 0.17 $\pm$ 0.04 & 0.59 $\pm$ 0.11 \\
 & MAC & 4.00 $\pm$ 0.82 & \textbf{0.14} $\pm$ 0.00 & \textbf{0.69} $\pm$ 0.02 \\
\bottomrule
\end{tabular}
\end{table*}


\begin{table*}[]
\centering
\caption{Comparison of methods on Child, Auto, Earthquake, Cancer, and Survey (with standard deviations) with other modules of MAC framework (DeepSeek-R1).}
\label{deepseek-combined}
\begin{tabular}{llccc}
\toprule
\textbf{Dataset} & \textbf{Method} & \textbf{SHD $(\downarrow)$} & \textbf{NHD $(\downarrow)$} & \textbf{F1 $(\uparrow)$} \\
\midrule
 & DCM & 30.67 $\pm$ 11.09 & 0.15 $\pm$ 0.00 & 0.19 $\pm$ 0.06 \\
Child & MDM & 23.00 $\pm$ 2.94 & 0.06 $\pm$ 0.01 & 0.48 $\pm$ 0.07 \\
 & MAC & \textbf{15.67} $\pm$ 3.09 & \textbf{0.05} $\pm$ 0.01 & \textbf{0.65} $\pm$ 0.07 \\
\midrule
 & DCM & \textbf{4.67} $\pm$ 1.25 & 0.35 $\pm$ 0.10 & 0.31 $\pm$ 0.14 \\
Auto & MDM & 7.33 $\pm$ 0.47 & 0.33 $\pm$ 0.02 & 0.24 $\pm$ 0.07 \\
 & MAC & 4.33 $\pm$ 0.94 & \textbf{0.21} $\pm$ 0.04 & \textbf{0.56} $\pm$ 0.08 \\
\midrule
 & DCM & 4.67 $\pm$ 0.47 & 0.32 $\pm$ 0.03 & 0.00 $\pm$ 0.00 \\
Earthquake & MDM & 0.33 $\pm$ 0.47 & 0.01 $\pm$ 0.02 & 0.96 $\pm$ 0.05 \\
 & MAC & \textbf{0.00} $\pm$ 0.00 & \textbf{0.00} $\pm$ 0.00 & \textbf{1.00} $\pm$ 0.00 \\
\midrule
 & DCM & 3.00 $\pm$ 0.82 & 0.17 $\pm$ 0.02 & 0.26 $\pm$ 0.20 \\
Cancer & MDM & 1.67 $\pm$ 0.94 & 0.07 $\pm$ 0.04 & 0.84 $\pm$ 0.08 \\
 & MAC & \textbf{1.00} $\pm$ 0.82 & \textbf{0.04} $\pm$ 0.03 & \textbf{0.90} $\pm$ 0.08 \\
\midrule
 & DCM & \textbf{2.00} $\pm$ 0.00 & 0.22 $\pm$ 0.08 & 0.25 $\pm$ 0.35 \\
Survey & MDM & 6.00 $\pm$ 0.82 & 0.17 $\pm$ 0.02 & 0.57 $\pm$ 0.08 \\
 & MAC & 4.00 $\pm$ 1.41 & \textbf{0.11} $\pm$ 0.04 & \textbf{0.71} $\pm$ 0.10 \\
\bottomrule
\end{tabular}
\end{table*}


\begin{table*}[]
\centering
\caption{Comparison of Methods on Child, Auto, Earthquake, Cancer, and Survey (with standard deviations) with other modules of MAC framework (GPT-4o).}
\label{gpt4o-individual-module-combined}
\begin{tabular}{llccc}
\toprule
\textbf{Dataset} & \textbf{Method} & \textbf{SHD $(\downarrow)$} & \textbf{NHD $(\downarrow)$} & \textbf{F1 $(\uparrow)$} \\
\midrule
 & DCM & 30.67 $\pm$ 11.09 & 0.15 $\pm$ 0.00 & 0.19 $\pm$ 0.06 \\
Child & MDM & 24.67 $\pm$ 2.87 & 0.07 $\pm$ 0.01 & 0.42 $\pm$ 0.10 \\
 & MAC & \textbf{20.33} $\pm$ 1.25 & \textbf{0.06} $\pm$ 0.01 & \textbf{0.41} $\pm$ 0.02 \\
\midrule
 & DCM & \textbf{4.67} $\pm$ 1.25 & 0.35 $\pm$ 0.10 & 0.31 $\pm$ 0.14 \\
Auto & MDM & 6.00 $\pm$ 0.00 & 0.27 $\pm$ 0.02 & 0.38 $\pm$ 0.02 \\
 & MAC & 5.00 $\pm$ 1.63 & \textbf{0.23} $\pm$ 0.05 & \textbf{0.43} $\pm$ 0.18 \\
\midrule
 & DCM & 4.67 $\pm$ 0.47 & 0.32 $\pm$ 0.03 & 0.00 $\pm$ 0.00 \\
Earthquake & MDM & \textbf{0.00} $\pm$ 0.00 & \textbf{0.00} $\pm$ 0.00 & \textbf{1.00} $\pm$ 0.00 \\
 & MAC & 0.33 $\pm$ 0.47 & 0.01 $\pm$ 0.02 & 0.96 $\pm$ 0.05 \\
\midrule
 & DCM & 3.00 $\pm$ 0.82 & 0.17 $\pm$ 0.02 & 0.26 $\pm$ 0.20 \\
Cancer & MDM & 2.00 $\pm$ 0.00 & 0.08 $\pm$ 0.00 & 0.80 $\pm$ 0.00 \\
 & MAC & \textbf{1.67} $\pm$ 1.70 & \textbf{0.07} $\pm$ 0.07 & \textbf{0.83} $\pm$ 0.17 \\
\midrule
 & DCM & \textbf{2.00} $\pm$ 0.00 & 0.22 $\pm$ 0.08 & 0.25 $\pm$ 0.35 \\
Survey & MDM & 3.33 $\pm$ 0.47 & 0.09 $\pm$ 0.01 & 0.74 $\pm$ 0.05 \\
 & MAC & 2.33 $\pm$ 0.47 & \textbf{0.06} $\pm$ 0.01 & \textbf{0.79} $\pm$ 0.01 \\
\bottomrule
\end{tabular}
\end{table*}

\begin{table*}[h]
\centering
\caption{Performance comparison of Single-Agent MAC vs.\ MAC (full multi-agent) across different LLMs. Lower is better for SHD and NHD; higher is better for F1.}
\label{tab:single_mac_performance}
\begin{tabular}{llccc}
\toprule
\textbf{Dataset} & \textbf{Method} & \textbf{SHD} (\(\downarrow\)) & \textbf{NHD} (\(\downarrow\)) & \textbf{F1} (\(\uparrow\)) \\
\midrule
\multicolumn{5}{l}{\textit{\textbf{Gemini-2.0-Flash}}} \\
\midrule
\multirow{2}{*}{Child}      & Single-Agent MAC & 25.33 $\pm$0.58 & 0.07 $\pm$0.01 & 0.15 $\pm$0.26 \\
                            & MAC              & 24.33 $\pm$3.09 & 0.07 $\pm$0.01 & 0.44 $\pm$0.09 \\
\addlinespace
\multirow{2}{*}{Auto}       & Single-Agent MAC & 5.33 $\pm$0.58  & 0.25 $\pm$0.09 & 0.06 $\pm$0.10 \\
                            & MAC              & 4.33 $\pm$0.47  & 0.19 $\pm$0.04 & 0.61 $\pm$0.08 \\
\addlinespace
\multirow{2}{*}{Earthquake} & Single-Agent MAC & 0.00 $\pm$0.00  & 0.00 $\pm$0.00 & 1.00 $\pm$0.00 \\
                            & MAC              & 0.00 $\pm$0.00  & 0.00 $\pm$0.00 & 1.00 $\pm$0.00 \\
\addlinespace
\multirow{2}{*}{Cancer}     & Single-Agent MAC & 1.67 $\pm$2.08  & 0.07 $\pm$0.08 & 0.63 $\pm$0.55 \\
                            & MAC              & 2.00 $\pm$0.00  & 0.08 $\pm$0.00 & 0.80 $\pm$0.00 \\
\addlinespace
\multirow{2}{*}{Survey}     & Single-Agent MAC & 9.00 $\pm$1.41  & 0.28 $\pm$0.04 & 0.45 $\pm$0.04 \\
                            & MAC              & 4.00 $\pm$0.82  & 0.14 $\pm$0.00 & 0.69 $\pm$0.02 \\
\midrule
\multicolumn{5}{l}{\textit{\textbf{DeepSeek-R1}}} \\
\midrule
\multirow{2}{*}{Child}      & Single-Agent MAC & 19.18 $\pm$3.98 & 0.05 $\pm$0.01 & 0.49 $\pm$0.21 \\
                            & MAC              & 15.67 $\pm$3.09 & 0.05 $\pm$0.01 & 0.65 $\pm$0.07 \\
\addlinespace
\multirow{2}{*}{Auto}       & Single-Agent MAC & 6.25 $\pm$1.50  & 0.35 $\pm$0.05 & 0.16 $\pm$0.12 \\
                            & MAC              & 4.33 $\pm$0.94  & 0.21 $\pm$0.04 & 0.56 $\pm$0.08 \\
\addlinespace
\multirow{2}{*}{Earthquake} & Single-Agent MAC & 0.25 $\pm$0.90  & 0.02 $\pm$0.07 & 0.95 $\pm$0.21 \\
                            & MAC              & 0.00 $\pm$0.00  & 0.00 $\pm$0.00 & 1.00 $\pm$0.00 \\
\addlinespace
\multirow{2}{*}{Cancer}     & Single-Agent MAC & 0.77 $\pm$1.51  & 0.03 $\pm$0.06 & 0.87 $\pm$0.30 \\
                            & MAC              & 1.00 $\pm$0.82  & 0.04 $\pm$0.03 & 0.90 $\pm$0.08 \\
\addlinespace
\multirow{2}{*}{Survey}     & Single-Agent MAC & 8.77 $\pm$1.30  & 0.26 $\pm$0.04 & 0.28 $\pm$0.12 \\
                            & MAC              & 4.00 $\pm$1.41  & 0.11 $\pm$0.04 & 0.71 $\pm$0.10 \\
\midrule
\multicolumn{5}{l}{\textit{\textbf{GPT-4o}}} \\
\midrule
\multirow{2}{*}{Child}      & Single-Agent MAC & 28.67 $\pm$1.53 & 0.09 $\pm$0.00 & 0.34 $\pm$0.01 \\
                            & MAC              & 20.33 $\pm$1.25 & 0.06 $\pm$0.01 & 0.41 $\pm$0.02 \\
\addlinespace
\multirow{2}{*}{Auto}       & Single-Agent MAC & 6.67 $\pm$1.15  & 0.41 $\pm$0.02 & 0.11 $\pm$0.10 \\
                            & MAC              & 5.00 $\pm$1.63  & 0.23 $\pm$0.05 & 0.43 $\pm$0.18 \\
\addlinespace
\multirow{2}{*}{Earthquake} & Single-Agent MAC & 0.00 $\pm$0.00  & 0.00 $\pm$0.00 & 1.00 $\pm$0.00 \\
                            & MAC              & 0.33 $\pm$0.47  & 0.01 $\pm$0.02 & 0.96 $\pm$0.05 \\
\addlinespace
\multirow{2}{*}{Cancer}     & Single-Agent MAC & 0.00 $\pm$0.00  & 0.00 $\pm$0.00 & 1.00 $\pm$0.00 \\
                            & MAC              & 1.67 $\pm$1.70  & 0.07 $\pm$0.07 & 0.83 $\pm$0.17 \\
\addlinespace
\multirow{2}{*}{Survey}     & Single-Agent MAC & 11.67 $\pm$0.58 & 0.33 $\pm$0.00 & 0.36 $\pm$0.04 \\
                            & MAC              & 2.33 $\pm$0.47  & 0.06 $\pm$0.01 & 0.79 $\pm$0.01 \\
\bottomrule
\end{tabular}
\end{table*}

\end{document}